\documentclass{article}

\usepackage{arxiv}

\usepackage[utf8]{inputenc} 
\usepackage[T1]{fontenc}    
\usepackage[hidelinks]{hyperref}       
\usepackage{url}            
\usepackage{booktabs}       
\usepackage{amsfonts}       
\usepackage{nicefrac}       
\usepackage{microtype}      
\usepackage{lipsum}         
\usepackage{graphicx}
\usepackage[numbers]{natbib}
\usepackage{doi}
\usepackage{nicefrac}
\usepackage{amsthm}
\usepackage[acronym]{glossaries}
\usepackage{subcaption}
\usepackage{booktabs} 
\usepackage{multirow}
\usepackage{fontawesome}
\usepackage{tikzit}
\usepackage{cleveref}       
\usepackage{wrapfig}
\usepackage{tcolorbox}

\usepackage{algorithm}
\usepackage{algorithmic}

\title{Multi-Source Domain Adaptation through Dataset Dictionary Learning in Wasserstein Space}

\date{}

\newif\ifuniqueAffiliation
\uniqueAffiliationtrue

\ifuniqueAffiliation 
\author{
Eduardo Fernandes Montesuma\\
CEA, List\\
Université Paris-Saclay\\
F-91120 Palaiseau, France
\And
Fred Ngolè Mboula\\
CEA, List\\
Université Paris-Saclay\\
F-91120 Palaiseau, France
\And
Antoine Souloumiac\\
CEA, List\\
Université Paris-Saclay\\
F-91120 Palaiseau, France
}
\else
\usepackage{authblk}

\newbox{\orcid}\sbox{\orcid}{\includegraphics[scale=0.06]{orcid.pdf}} 
\author[1]{%
	\href{https://orcid.org/0000-0000-0000-0000}{\usebox{\orcid}\hspace{1mm}David S.~Hippocampus\thanks{\texttt{hippo@cs.cranberry-lemon.edu}}}%
}
\author[1,2]{%
	\href{https://orcid.org/0000-0000-0000-0000}{\usebox{\orcid}\hspace{1mm}Elias D.~Striatum\thanks{\texttt{stariate@ee.mount-sheikh.edu}}}%
}
\affil[1]{Department of Computer Science, Cranberry-Lemon University, Pittsburgh, PA 15213}
\affil[2]{Department of Electrical Engineering, Mount-Sheikh University, Santa Narimana, Levand}
\fi

\renewcommand{\headeright}{Dataset Dictionary Learning}
\renewcommand{\undertitle}{}
\renewcommand{\shorttitle}{}

\hypersetup{
pdftitle={Dataset Dictionary Learning},
pdfsubject={cs.AI},
pdfauthor={Eduardo Fernandes Montesuma,Fred Ngol\'e Mboula,Antoine Souloumiac},
pdfkeywords={Optimal Transport,Transfer Learning,Domain Adaptation},
}

\newacronym{ot}{OT}{Optimal Transport}
\newacronym{dil}{DiL}{Dictionary Learning}
\newacronym{ml}{ML}{Machine Learning}
\newacronym{erm}{ERM}{Empirical Risk Minimization}
\newacronym{da}{DA}{Domain Adaptation}
\newacronym{tl}{TL}{Transfer Learning}
\newacronym{msda}{MSDA}{Multi-Source DA}
\newacronym{nmf}{NMF}{Non-negative Matrix Factorization}
\newacronym{sota}{SOTA}{State-of-the-Art}
\newacronym{wdl}{WDL}{Wasserstein Dictionary Learning}
\newacronym{dadil}{DaDiL}{Dataset Dictionary Learning}

\newacronym{tca}{TCA}{Transfer Component Analysis}
\newacronym{otda}{OTDA}{Optimal Transport Domain Adaptation}
\newacronym{sa}{SA}{Subspace Alignment}
\newacronym{coral}{CORAL}{Correlation Alignment}
\newacronym{jdot}{JDOT}{Joint Distribution Optimal Transport}
\newacronym{wbt}{WBT}{Wasserstein Barycenter Transport}

\newacronym{dann}{DANN}{Domain Adversarial Neural Network}
\newacronym{wdgrl}{WDGRL}{Wasserstein Distance Guided Representation Learning}
\newacronym{mcd}{MCD}{Maximum Classifier Discrepancy}
\newacronym{mdd}{MDD}{Margin Disparity Discrepancy}
\newacronym{wjdot}{WJDOT}{Weighted JDOT}
\newacronym{m3sda}{M3SDA}{Moment Matching for MSDA}

\newacronym{tsne}{t-SNE}{t-Stochastic Neighbor Embeddings}
\newacronym{nn}{NN}{Neural Net}
\newacronym{pot}{POT}{Python Optimal Transport}
\newacronym{wb}{WB}{Wasserstein Barycenter}
\newacronym{wbr}{WBR}{Wasserstein Barycentric Coordinates Regression}
\newacronym{dnn}{DNN}{Deep Neural Net}

\tikzstyle{trapezium}=[fill=white, draw=black, shape=trapezium, rotate=-90, minimum height=1cm]
\tikzstyle{lossbox}=[fill={rgb,255: red,202; green,206; blue,255}, draw=black, shape=rectangle, minimum height=1.2cm, minimum width=1cm, align=center]
\tikzstyle{clfbox}=[fill=white, draw=black, shape=rectangle, minimum width=1cm, minimum height=1cm]
\tikzstyle{new style 2}=[fill=white, draw=black, shape=rectangle, align=center]
\tikzstyle{domainbox}=[fill=white, draw=black, shape=rectangle, minimum width=3cm, align=center]
\tikzstyle{longbox}=[fill=white, draw=black, shape=rectangle, minimum height=4cm, minimum width=1.2cm, align=center]
\tikzstyle{rotatednode}=[rotate=90]
\tikzstyle{circularnode}=[fill={rgb,255: red,204; green,204; blue,204}, draw=black, shape=circle]
\tikzstyle{blue_circle}=[fill={rgb,255: red,0; green,80; blue,104}, draw=none, shape=circle, minimum width=0.5cm]
\tikzstyle{orangecircle1}=[fill={rgb,255: red,231; green,111; blue,81}, draw=none, shape=circle, minimum width=0.5cm]
\tikzstyle{blue_square1}=[fill={rgb,255: red,0; green,80; blue,104}, draw=none, shape=rectangle, minimum width=0.5cm, minimum height=0.5cm]
\tikzstyle{blue_triangle1}=[fill={rgb,255: red,0; green,80; blue,104}, draw=none, shape=regular polygon, regular polygon sides=3]
\tikzstyle{widebox}=[fill=white, draw=black, shape=rectangle, minimum height=1.2cm, minimum width=10cm, align=center]
\tikzstyle{labeled domain}=[fill=none, draw={rgb,255: red,0; green,80; blue,104}, shape=circle, minimum width=1cm]
\tikzstyle{unlabeled domain}=[fill=none, draw={rgb,255: red,231; green,111; blue,81}, shape=circle, minimum width=1cm]

\tikzstyle{red edge}=[->, fill=none, draw={rgb,255: red,128; green,0; blue,0}]
\tikzstyle{blue edge}=[->, fill=none, draw={rgb,255: red,70; green,130; blue,180}]
\tikzstyle{green edge}=[->, fill=none, draw={rgb,255: red,44; green,160; blue,44}]
\tikzstyle{red dotted edge}=[->, dashed, fill=none, draw={rgb,255: red,128; green,0; blue,0}]
\tikzstyle{blue dotted edge}=[->, dashed, fill=none, draw={rgb,255: red,70; green,130; blue,180}]
\tikzstyle{green dotted edge}=[->, dashed, fill=none, draw={rgb,255: red,44; green,160; blue,44}]
\tikzstyle{red dotted line}=[-, fill=none, dashed, draw={rgb,255: red,128; green,0; blue,0}]
\tikzstyle{blue dotted line}=[-, fill=none, dashed, draw={rgb,255: red,70; green,130; blue,180}]
\tikzstyle{green dotted line}=[-, fill=none, dashed, draw={rgb,255: red,44; green,160; blue,44}]
\tikzstyle{black edge}=[->]
\tikzstyle{black dashed line}=[-, dashed]
\tikzstyle{thick black arrow}=[->, thick]
\tikzstyle{thick black edge}=[-, thick]

\renewcommand{\inf}[1]{\underset{#1}{\text{inf}}\,}

\newcommand{\iid}[0]{\overset{i.i.d.}{\sim}}
\newcommand{\argmin}[1]{\underset{#1}{\text{argmin}}\,}

\newcommand{\expectation}[1]{\underset{#1}{\mathbb{E}}}

\newtheorem{definition}{Definition}
\newtheorem{lemma}{Lemma}

\newtheorem{theorem}{Theorem}

\begin{document}
\maketitle

\begin{abstract}
This paper seeks to solve Multi-Source Domain Adaptation (MSDA), which aims to mitigate data distribution shifts when transferring knowledge from multiple labeled source domains to an unlabeled target domain. We propose a novel MSDA framework based on dictionary learning and optimal transport. We interpret each domain in MSDA as an empirical distribution. As such, we express each domain as a Wasserstein barycenter of dictionary atoms, which are empirical distributions. We propose a novel algorithm, DaDiL, for learning via mini-batches: (i) atom distributions; (ii) a matrix of barycentric coordinates. Based on our dictionary, we propose two novel methods for MSDA: DaDil-R, based on the reconstruction of labeled samples in the target domain, and DaDiL-E, based on the ensembling of classifiers learned on atom distributions. We evaluate our methods in 3 benchmarks: Caltech-Office, Office 31, and CRWU, where we improved previous state-of-the-art by 3.15\%, 2.29\%, and 7.71\% in classification performance. Finally, we show that interpolations in the Wasserstein hull of learned atoms provide data that can generalize to the target domain\footnote{Accepted as a conference paper at the 26th European Conference on Artificial Intelligence}.
\begin{tcolorbox}
\begin{center}
\faGithub\,\,\,\url{https://github.com/eddardd/demo-dadil/}
\end{center}
\end{tcolorbox}
\end{abstract}


\section{Introduction}\label{sec:intro}

Traditional \gls{ml} works under the assumption that training and test data follow a single probability distribution. Indeed, the \gls{erm} framework of [37] measures generalization regarding an unknown probability distribution from which training and test data are sampled. Nonetheless, as [25] remarks, this is seldom the case in realistic applications due to changes in how the data is acquired. This results in a change in the data distribution, or distributional shift that motivates the field of \gls{da}.

\gls{da} is an important framework where one assumes labeled data from a source domain and seeks to adapt models to an unlabeled target domain. When multiple source domains are available, one has a \gls{msda} setting. This problem is more challenging as one has multiple distributional shifts co-occurring, that is, between sources and between sources and the target. In this work, we assume that the domain shifts have regularities that can be learned and leveraged for \gls{msda}. In this context, \gls{ot} is a mathematical theory useful for \gls{da}, as it allows for the comparison and matching probability distributions. Previous works employed \gls{ot} for the single-source case, as in~\cite{courty2016optimal,courty2017joint,damodaran2018deepjdot}, and \gls{msda} as in~\cite{montesuma2021cvpr,montesuma2021icassp,turrisi2022multi}.

In parallel, \gls{dil} expresses a set of vectors as weighted combinations of dictionary elements, named atoms. Previous works proposed \gls{ot} for \gls{dil} over histogram data, such as \cite{rolet2016fast} and \cite{schmitz2018wasserstein}. Nonetheless, when data is high-dimensional, modeling distributions as histograms is intractable due the curse of dimensionality, which limits the use of previous \gls{dil} works for \gls{msda}.

\noindent\textbf{Contributions.} In this paper we propose a novel \gls{dil} framework  (section~\ref{sec:dadil}), for distributions represented as point clouds. We further explore (section~\ref{sec:dadil_msda}) two ways of using \gls{dil} for \gls{msda}, by reconstructing labeled samples in the target domain, and by ensembling classifiers learned with labeled data from atoms. In addition, we justify these methods theoretically through results in the literature~\cite[Theorem 2]{redko2017theoretical}, and through novel theoretical results (i.e., theorem~\ref{thm:dadil_guarantee}). To the best of our knowledge this is the first work to propose a \gls{dil} of point clouds, and to explore the connections between \gls{dil} of distributions and \gls{msda}.

\noindent\textbf{Paper Organization.} Section~\ref{sec:related_Work} covers the related literature. Section~\ref{sec:background} covers the necessary background, i.e., \gls{da}, \gls{ot} and \gls{dil} concepts. Section~\ref{sec:dadil} presents our framework. Section~\ref{sec:experiments} explores our experiments in \gls{msda}. Section~\ref{sec:discussion} discusses our results. Finally, section~\ref{sec:conclusion} concludes our paper.

\section{Related Work}\label{sec:related_Work}

There are mainly two methodologies in \gls{da}. The first, shallow \gls{da}, leverages pre-existing feature extractors and performs adaptation either by re-weighting or transforming source domain data to resemble target domain data. The second, deep \gls{da}, uses source and target domain data during the training of a \gls{dnn}, so that learned features are independent of distributional shift.

There are at least 3 classes of shallow \gls{da} methods: (i) importance re-weighting strategies~\cite{sugiyama2007covariate}, which give importance to source samples similar to the target domain, (ii) projection-based methods~\cite{pan2009survey}, which seek a sub-space where distributions share common characteristics, and (iii) \gls{ot}-based methods~\cite{courty2016optimal}, which use \gls{ot} for matching, or calculating distances between distributions. 

For deep \gls{da}, methods penalize encoder parameters that map source and target distributions to different locations in the latent space. As a consequence, deep \gls{da} is more complex than shallow \gls{da}, since encoder parameters are free. Examples of deep \gls{da} methods include~\cite{ganin2016domain}, who uses an adversarial loss, and~\cite{damodaran2018deepjdot, shen2018wasserstein}, who use \gls{ot} as a loss function between distributions of latent representations.

For \gls{msda}, some works generalize previous single-source baselines. For instance,~\cite{peng2019moment} proposes a moment-matching strategy across the different domains.~\cite{turrisi2022multi} proposes weighting source domains linearly, then applying the \gls{jdot} strategy of~\cite{courty2017joint}. This approach combines notions of importance weighting and \gls{ot}-based \gls{da}.~\cite{montesuma2021cvpr,montesuma2021icassp} generalize the approach of~\cite{courty2016optimal}, by first calculating a Wasserstein barycenter of the different source domains, then transporting the barycenter to the target domain.

In parallel, \gls{dil} is a representation learning technique, that was previously used in \gls{da} by~\cite{huang2013coupled}. However, classic \gls{dil} lacks a probabilistic interpretation. In this context, \gls{ot} offers a probabilistic foreground for \gls{dil}, when data is represented through histograms. This is done by either using the Sinkhorn divergence~\cite{cuturi2013sinkhorn} as the objective function~\cite{rolet2016fast}, or by aggregating atoms in a Wasserstein space~\cite{schmitz2018wasserstein}. Nonetheless, in the context of \gls{da} it is computationally intractable to bin the feature space, which is commonly high-dimensional. This issue hinders the applicability of previous \gls{dil} approaches for \gls{da}. In contrast, we propose a new \gls{ot}-inspired \gls{dil} framework for point clouds, which makes it suitable for \gls{msda}.

\section{Background}\label{sec:background}
\subsection{Domain Adaptation}

In \gls{ml}, learning a classifier consists on estimating $h:\mathcal{X}\rightarrow\mathcal{Y}$, among a set of functions $\mathcal{H}$, where $\mathcal{X}$ (e.g., $\mathbb{R}^{d}$) is the \emph{feature space} and $\mathcal{Y}$ (e.g., $\{1,\cdots,n_{c}\}$) is the \emph{label space}. This estimation is done via \emph{risk minimization}~\cite{vapnik1991principles},
\begin{align}
    h^{\star} = \argmin{h\in\mathcal{H}}\mathcal{R}_{Q}(h)=\expectation{\mathbf{x}\sim Q}[\mathcal{L}(h(\mathbf{x}),h_{0}(\mathbf{x}))],\label{eq:err}
\end{align}
for a loss function $\mathcal{L}$, a distribution $Q$, and a ground-truth labeling function $h_{0}$. $\mathcal{R}_{Q}$ is known as \emph{true risk}. Since $Q$ and $h_{0}$ are seldom known \emph{a priori}, it is unfeasible to directly minimize equation~\ref{eq:err}. In practice, one acquires a \emph{dataset}, $\{(\mathbf{x}_{i}^{(Q)},y_{i}^{(Q)})\}_{i=1}^{n}$, with $\mathbf{x}_{i}^{(Q)} \overset{\text{i.i.d.}}{\sim} Q$ and $y_{i}^{(Q)} = h_{0}(\mathbf{x}_{i}^{(Q)})$ and minimizes the \emph{empirical risk},
\begin{align*}
    \hat{h} = \argmin{h\in\mathcal{H}}\hat{\mathcal{R}}_{Q}(h)=\dfrac{1}{n}\sum_{i=1}^{n}\mathcal{L}(h(\mathbf{x}_{i}^{(Q)}),y_{i}^{(Q)}).
\end{align*}
Henceforth $\mathbf{x}_{i}^{(Q)}$ denotes a feature vector sampled from the marginal $Q(X)$. Likewise, $y_{i}^{(Q)}$ denotes its corresponding label. We denote its corresponding one-hot encoding (hard-labels) or probability vector (soft-labels) by $\mathbf{y}_{i}^{(Q)}$.

As discussed in~\cite{redko2020survey}, if training and test data are i.i.d. from $Q$, $\mathcal{R}_{Q} \rightarrow \hat{\mathcal{R}}_{Q}$ as $n\rightarrow \infty$. Nonetheless this assumption is restrictive, as it disregards the distributional heterogeneity within training data, and between train and test data, which motivates \gls{da}~\cite{pan2009survey}. Following~\cite{pan2009survey}, a domain $\mathcal{D}=(\mathcal{X},Q(X))$ is a pair of a feature space, and a feature distribution. In \gls{da}, one has different domains, i.e., a labeled source $\mathcal{D}_{S}$ with samples $\{(\mathbf{x}_{i}^{(Q_{S})},y_{i}^{(Q_{S})})\}_{i=1}^{n_{Q_{S}}}$ and a target $\mathcal{D}_{T}$ with samples $\{\mathbf{x}_{j}^{(Q_{T})}\}_{j=1}^{n_{Q_{T}}}$. In practice, one assumes a shared feature space (e.g., $\mathbb{R}^{d}$), so that domains differ in their distribution, $Q_{S}(X) \neq Q_{T}(X)$. This is known in the literature as \emph{distributional shift}. The goal of \gls{da} is improving performance on the target, given knowledge from the source domain. We investigate \gls{msda}, that is, \gls{da} between labeled sources $\{\mathcal{D}_{S_{\ell}}\}_{\ell=1}^{N_{S}}$ and an unlabeled target $\mathcal{D}_{T}$. 



\subsection{Optimal Transport}

\gls{ot} is a field of mathematics widely used in \gls{da} and \gls{ml}. Henceforth we focus on computational \gls{ot}. We refer readers to~\cite{montesuma2023recent} and~\cite{peyre2019computational} for further background on this topic. Let $\mathbf{x}_{i}^{(P)} \iid P$ (resp. $\mathbf{x}_{j}^{(Q)} \iid Q$). We $P$ and $Q$ empirically using mixtures of Dirac deltas,
\begin{equation}
    \hat{P}(\mathbf{x}) = \dfrac{1}{n_{P}}\sum_{i=1}^{n_{P}}\delta(\mathbf{x}-\mathbf{x}_{i}^{(P)}).\label{eq:empirical_approx}
\end{equation}
We refer to $\hat{P}$ as a point cloud, and $\mathbf{X}^{(P)} = [\mathbf{x}_{1}^{(P)},\cdots,\mathbf{x}_{n_{P}}^{(P)}] \in \mathbb{R}^{n_{P} \times d}$ to its \emph{support}. The Kantorovich formulation of \gls{ot} seeks an \gls{ot} plan, $\pi \in \mathbb{R}^{n_{P} \times n_{Q}}$ that \emph{preserves mass},
\begin{equation*}
    \Pi(\hat{P},\hat{Q}) := \{\pi:\sum_{i}\pi_{i,j}=\nicefrac{1}{n_{Q}};\sum_{j}\pi_{i,j}=\nicefrac{1}{n_{P}}\}.
\end{equation*}
where $\pi_{i,j}$ denotes how much mass $\mathbf{x}_{i}^{(P)}$ sends to $\mathbf{x}_{j}^{(Q)}$. In this sense, the \gls{ot} problem between $\hat{P}$ and $\hat{Q}$ is,
\begin{equation}
    \pi^{\star} = \text{OT}(\mathbf{X}^{(P)},\mathbf{X}^{(Q)}) = \argmin{\pi\in \Pi(\hat{P},\hat{Q})}\langle \mathbf{C},\pi \rangle_{F},\label{eq:KantorovichPr}
\end{equation}
where $\langle\cdot,\cdot\rangle_{F}$ denotes the Frobenius inner product and $C_{i,j}=c(\mathbf{x}_{i}^{(P)},\mathbf{x}_{j}^{(Q)})$ is called \emph{ground-cost matrix}. This is a linear program on the variables $\pi_{i,j}$, which has computational complexity $\mathcal{O}(n^{3}\log n)$. Given $\pi$, one often wants to map samples from $P$ into $Q$, which can be done through the \emph{barycentric projection}~\cite{courty2016optimal},
\begin{equation*}
    T_{\pi}(\mathbf{x}_{i}^{(P)}) = \argmin{\mathbf{x}\in\mathbb{R}^{d}}\sum_{j=1}^{n_{Q}}\pi_{i,j}c(\mathbf{x},\mathbf{x}_{j}^{(Q)}).
\end{equation*}
When $c$ is the Euclidean distance, the barycentric projection has closed form, 
\begin{equation}
    T_{\pi}(\mathbf{x}_{i}^{(P)}) = n_{P}\sum_{j=1}^{n_{Q}}\pi_{i,j}\mathbf{x}_{j}^{(Q)},\label{eq:barymap}
\end{equation}
or $T_{\pi}(\mathbf{X}^{(P)}) = n_{P}\pi\mathbf{X}^{(Q)}$ in short.

\noindent\textbf{Optimal Transport for Domain Adaptation.} In the semminal works of~\cite{courty2016optimal}, the authors proposed using \gls{ot} for \gls{da}, under the assumption that there is $T:\mathbb{R}^{d}\rightarrow\mathbb{R}^{d}$ such that,
\begin{align}
T_{\sharp}Q_{S} = Q_{T}\text{ and }Q_{S}(Y|X)=Q_{T}(Y|T(X)),\label{eq:hypothesis_courty}
\end{align}
where $T_{\sharp}$ is the push-forward operator (see e.g.,~\cite{santambrogio2015optimal}).~\cite{courty2016optimal} propose estimating $T$ through $T_{\pi}$ in equation~\ref{eq:barymap}, which allows mapping samples from $Q_{S}$ to $Q_{T}$.

\noindent\textbf{Wasserstein Barycenters.} When the ground-cost is a distance, \gls{ot} defines a distance between distributions, $W_{c}(\hat{P},\hat{Q}) = \langle \mathbf{C}, \pi^{\star} \rangle_{F}$, called Wasserstein distance. As such, \gls{ot} defines barycenters of probability distributions~\cite{agueh2011barycenters}. Henceforth we denote the $K-$simplex as $\Delta_{K} = \{\mathbf{a} \in \mathbb{R}^{K}_{+}:\sum_{k}a_{k}=1\}$.

\begin{definition}\label{def:wbary}
For distributions $\mathcal{P} = \{P_{k}\}_{k=1}^{K}$ and weights $\alpha \in \Delta_{K}$, the Wasserstein barycenter is a solution to,
\begin{align}
    B^{\star} = \mathcal{B}(\alpha;\mathcal{P}) = \inf{B}\sum_{k=1}^{K}\alpha_{k}W_{c}(P_{k}, B).\label{eq:true_bary}
\end{align}
Henceforth we call $\mathcal{B}(\cdot;\mathcal{P})$ barycentric operator. In this context, the Wasserstein hull of distributions $\mathcal{P}$ is,
\begin{align}
    \mathcal{M}(\mathcal{P}) = \{\mathcal{B}(\alpha; \mathcal{P}):\alpha \in \Delta_{K}\}
\end{align}
\end{definition}

When the distributions in $\mathcal{P}$ are empirical, solving equation~\ref{eq:true_bary} corresponds to estimating the support $\mathbf{X}^{(B)}$ of $B$. In this context,~\cite{cuturi2014fast} proposed an algorithm known as \emph{free-support Wasserstein barycenter} for calculating $\hat{B}$. Let $\mathbf{x}_{i}^{(B)} \sim \mathcal{N}(\mathbf{0},\mathbf{I}_{d})$ be an initialization for the barycenter's support. One updates the support of $\hat{B}$ with,
\begin{equation}
    \begin{aligned}
        \pi^{(k,it)} &= \text{OT}(\mathbf{X}^{(P_{k})}, \mathbf{X}^{(B)}_{it})\\
        \mathbf{X}^{(B)}_{it+1} &\leftarrow \theta\mathbf{X}^{(B)}_{it} + (1-\theta)\sum_{k=1}^{K}\alpha_{k}T_{\pi^{(k,it)}}(\mathbf{X}^{(B)}_{it})
    \end{aligned}\label{eq:iterations_wbary}
\end{equation}
where $\theta$ is found at each iteration via line search. In the context of \gls{msda},~\cite{montesuma2021cvpr} previously defined a barycenter of labeled distributions by penalizing transport plans $\pi^{(k, it)}$ that mix classes.

\noindent\textbf{Mini-batch OT.} For large scale datasets, computing \gls{ot} is likely unfeasible, due its cubic complexity. A workaround, coming from \gls{ml}, consists on using mini-batches~\cite{fatras2021minibatch}. For $M$ batches of size $n_{b}$, this approach decreases the time complexity to $\mathcal{O}(Mn_{b}^{3}\log n_{b})$.

\noindent\textbf{Remark on Notation.} While $W_{2}(\hat{P},\hat{Q})$ is defined between \emph{empirical distributions}, in practice it is a function of $(\mathbf{X}^{(P)},\mathbf{X}^{(Q)})$. With an abuse of notation, the mini-batch Wasserstein distance between given random samples of size $n_{b}$ from $P$ and $Q$ is still noted as $W_{2}(\hat{P}, \hat{Q})$, with the support matrices restricted to a mini-batch.

\subsection{Dictionary Learning}

\gls{dil} is a representation learning technique that expresses a collection of vectors $\{\mathbf{x}_{\ell}\}_{\ell=1}^{N}$, $\mathbf{x}_{\ell} \in \mathbb{R}^{d}$ through a set of atoms $\mathcal{P} = \{\mathbf{p}_{k}\}_{k=1}^{K}$, $\mathbf{p}_{k} \in \mathbb{R}^{d}$ and weights $\mathcal{A} = \{\alpha_{\ell}\}_{\ell=1}^{N}$, $\alpha_{\ell} \in \mathbb{R}^{K}$. Mathematically,
\begin{align*}
    \argmin{\mathcal{P},\mathcal{A}}\dfrac{1}{N}\sum_{i=1}^{N}\mathcal{L}(\mathbf{x}_{\ell}, \mathcal{P}^{T}\alpha_{\ell})+\lambda_{A}\Omega_{A}(\mathcal{A})+\lambda_{P}\Omega_{P}(\mathcal{P}),
\end{align*}
where $\mathcal{L}$ is a suitable loss, $\Omega_{A}$ and $\Omega_{P}$ are regularizing terms on $\mathcal{A}$ and $\mathcal{P}$ respectively. In this sense, \gls{ot} has previously contributed to \gls{dil} either by defining a meaningful loss function, or novel ways to aggregating atoms. For instance,~\cite{rolet2016fast} proposed using the Sinkhorn divergence of~\cite{cuturi2013sinkhorn} as a loss function, while~\cite{schmitz2018wasserstein} proposed using Wasserstein barycenters for aggregating atoms. These works assume data in the form of histograms, i.e., $\mathbf{x}_{\ell} \in \Delta_{d}$. As consequence, $\mathbf{p}_{k} \in \Delta_{d}$ and $\alpha_{\ell} \in \Delta_{K}$.

\section{Proposed Framework}\label{sec:dadil}

In this section, we present our novel framework for \gls{msda}, called \gls{dadil}. As our discussion relies on analogies with \gls{dil} theory, we provide in Table~\ref{tab:analogies} a comparison of \gls{dil} concepts in different frameworks. In what follows, section~\ref{sec:labeled_wbary} presents a novel algorithm for computing Wasserstein barycenters of labeled distributions, and section~\ref{sec:dadil_msda} presents our framework.

\begin{table}[ht]
    \centering
    \caption{Overview of analogies between different frameworks of \gls{dil}.}
    \begin{tabular}{ccccc}
         \toprule
        Concept & Symbol & Classic DiL & WDL~\citep{schmitz2018wasserstein} & DaDiL (ours) \\
         \midrule
        Data & $\mathbf{x}_{\ell}$, or $\hat{Q}_{\ell}$ & Vectors & Histograms & Point Clouds\\
        Atom & $\mathcal{P}$ & Vectors & Histograms & Point Clouds \\
        Representation & $\mathcal{A}$ & Vectors & Barycentric Coordinates & Barycentric Coordinates \\
        Reconstruction & $\mathcal{B}$ & Vectors & Histograms & Point Clouds \\
         \bottomrule
    \end{tabular}
    \label{tab:analogies}
\end{table}

\subsection{Wasserstein Barycenters of Labeled Distributions}\label{sec:labeled_wbary}

We propose a novel algorithm for calculating differentiable Wasserstein barycenters of labeled empirical distributions. This algorithm is at the core of \gls{dadil} (section~\ref{sec:dadil}), since we later represent datasets as barycenters of learned atoms.

In \gls{ot}, there are at least 2 ways of integrating labels, either by penalizing \gls{ot} plans that transport mass between different classes~\citep{courty2016optimal,montesuma2021cvpr}, or by defining a metric in the label space~\citep{alvarez2020geometric}. We choose to integrate labels in the ground-cost,
\begin{equation}
    C_{i,j} = \lVert \mathbf{x}_{i}^{(P)} - \mathbf{x}_{j}^{(Q)} \rVert_{2}^{2} + \beta \lVert \mathbf{y}^{(P)}_{i}-\mathbf{y}^{(Q)}_{j}\rVert_{2}^{2},\label{eq:supervised_ground_cost}
\end{equation}
\begin{wrapfigure}{r}{0.5\textwidth}
    \begin{minipage}{0.5\textwidth}
        \begin{algorithm}[H]
        \caption{Labeled Wasserstein Barycenter}
        \begin{algorithmic}[1]
            \REQUIRE $\{\mathbf{X}^{(P_{k})},\mathbf{Y}^{(P_{k})}\}_{k=1}^{K}$, $\alpha \in \Delta_{K}$, $\tau > 0$, $N_{itb}$.
            \FOR{$i=1,\cdots, n_{B}$}
                \STATE $\mathbf{x}^{(B)}_{i} \sim \mathcal{N}(\mathbf{0},\mathbf{I}_{d})$, $y_{i}^{(B)} = \text{randint}(n_{c})$
            \ENDFOR
            \WHILE{$|J_{it} - J_{it-1}| \geq \tau$ and $it \leq N_{itb}$}
                \FOR{$k=1,\cdots K$}
                    \STATE $\pi^{(k,it)} = \text{OT}\biggr((\mathbf{X}^{(P_{k})}, \mathbf{Y}^{(P_{k})});(\mathbf{X}^{(B)}_{it}, \mathbf{Y}^{(B)}_{it})\biggr)$
                \ENDFOR
                \STATE $J_{it} = \sum_{k=1}^{K}\alpha_{k}\langle \pi^{(k,it)},\mathbf{C}^{(k)} \rangle_{F}$
                \STATE $\mathbf{X}^{(B)}_{it+1} = \sum_{k=1}^{K}\alpha_{k}T_{\pi^{(k,it)}}(\mathbf{X}^{(B)}_{it})$
                \STATE $\mathbf{Y}^{(B)}_{it+1} = \sum_{k=1}^{K}\alpha_{k}T_{\pi^{(k,it)}}(\mathbf{Y}^{(B)}_{it})$
            \ENDWHILE
            \ENSURE Labeled barycenter support $(\mathbf{X}^{(B)},\mathbf{Y}^{(B)})$.
        \end{algorithmic}
        \label{alg:wbt_fixed_point_joint}
        \end{algorithm}
    \end{minipage}
    \vspace{-20pt}
\end{wrapfigure}
\noindent where $\mathbf{y}$ denotes labels one-hot encoding, and $\beta > 0$ controls the importance of label discrepancy. While simple, this choice allows us to motivate the barycentric projection of~\cite{courty2016optimal}, and the label propagation of~\cite{redko2019advances} as first-order optimality conditions of $W_{c}(\hat{P}, \hat{Q})$,
\begin{equation}
    \begin{cases}
        \hat{\mathbf{x}}^{(P)}_{i} = T_{\pi}(\mathbf{x}_{i}^{(P)}) = n_{P}\sum_{j=1}^{n_{Q}}\pi_{i,j}\mathbf{x}_{j}^{(Q)},\\
        \hat{\mathbf{y}}^{(P)}_{i} = T_{\pi}(\mathbf{y}_{i}^{(P)}) = n_{P}\sum_{j=1}^{n_{Q}}\pi_{i,j}\mathbf{y}_{j}^{(Q)}.
    \end{cases}\label{eq:supervised_barycenter}
\end{equation}
Henceforth we denote,
\begin{equation*}
    \pi = \text{OT}\biggr((\mathbf{X}^{(P)},\mathbf{Y}^{(P)});(\mathbf{X}^{(Q)},\mathbf{Y}^{(Q)})\biggr).
\end{equation*}
As a consequence, we can interpolate between two point clouds, since $\hat{\mathbf{y}}_{i}^{(P)}$ corresponds to a soft-label (i.e., probabilities). We use equations~\ref{eq:supervised_ground_cost} and~\ref{eq:supervised_barycenter} for proposing a new barycenter strategy between labeled point clouds, shown in algorithm~\ref{alg:wbt_fixed_point_joint}.

\noindent\textbf{Differentiation.} For calculating derivatives of $\mathbf{x}_{i}^{(B)}$ and $\mathbf{y}_{i}^{(B)}$ w.r.t. $\mathbf{x}_{l}^{(P_{k})}$, $\mathbf{y}_{l}^{(P_{k})}$, and $\alpha$, we use the Envelope theorem of~\cite{afriat1971theory}. In other words, we do not propagate derivatives through the iterations of algorithm~\ref{alg:wbt_fixed_point_joint}. We provide further details in our appendix.

\noindent\textbf{Computational Complexity.} Let $\hat{P}_{k}$ have $n$ points in its support, for $k=1,\cdots,K$. The complexity of algorithm~\ref{alg:wbt_fixed_point_joint} is dominated by line 6, which has complexity $\mathcal{O}(n^{3}\log n)$. Hence, the overall computational complexity is $\mathcal{O}(N_{itb}Kn^{3} \log n)$.


\subsection{Dataset Dictionary Learning for MSDA}\label{sec:dadil_msda}

In this section, we introduce our novel framework, called \gls{dadil}, and explore how to use it for \gls{msda}. Let $\mathcal{Q} = \{\hat{Q}_{S_{\ell}}\}_{\ell=1}^{N_{S}} \cup \{\hat{Q}_{T}\}$ correspond to $N_{S}$ labeled sources and an unlabeled target. Let $\mathcal{A} = [\alpha_{1},\cdots,\alpha_{N_{S}},\alpha_{N_{S}+1}]$, and $\mathcal{P} = \{\hat{P}_{k}\}_{k=1}^{K}$. The $\hat{P}_{k}$'s are an empirical approximation of the point clouds that interpolate distributional shift. Following our notation, $\alpha_{T} := \alpha_{N_{S}+1}$. For $N = N_{S} + 1$, \gls{dadil} consists on minimizing,
\begin{align}
    (\mathcal{P}^{\star},\mathcal{A}^{\star}) = \argmin{\mathcal{P},\mathcal{A}\in(\Delta_{K})^{N}}\dfrac{1}{N}\sum_{\ell=1}^{N}\mathcal{L}(\hat{Q}_{\ell}, \mathcal{B}(\alpha_{\ell};\mathcal{P})),\label{eq:dadil}
\end{align}



\noindent where $\mathcal{L}$ is a loss between distributions. Since the target domain is not labeled, we define,
\begin{align*}
    \mathcal{L}(\hat{Q}_{\ell},\hat{B}_{\ell}) = \begin{cases}
        W_{c}(\hat{Q}_{\ell},\hat{B}_{\ell}),&\text{ if $\hat{Q}_{\ell}$ is labeled,}\\
        W_{2}(\hat{Q}_{\ell},\hat{B}_{\ell}),&\text{ otherwise,}
    \end{cases}
\end{align*}
i.e., when no labels in $\hat{Q}_{\ell}$ are available, we minimize the standard 2-Wasserstein distance. Optimizing~\ref{eq:dadil} over entire datasets might be intractable due the complexity of \gls{ot}. We thus employ mini-batch \gls{ot}~\cite{fatras2021minibatch}. In addition, we need to enforce the constraints $\mathbf{y}^{(P_{k})}_{l} \in \Delta_{n_{c}}$ and $\alpha_{\ell} \in \Delta_{K}$. In the first case we do a change of variables, and optimize the logits $\mathbf{p}\in\mathbb{R}^{n_{c}}\text{ s.t. }\mathbf{y} = \text{softmax}(\mathbf{p})$. In the second case, we project $\alpha_{\ell}$ into the simplex orthogonally,
\begin{align*}
    \text{proj}_{\Delta_{K}}(\alpha_{\ell}) = \argmin{\alpha \in \Delta_{K}}\lVert 
\alpha-\alpha_{\ell} \rVert_{2}.
\end{align*}
The overall optimization algorithm is shown in algorithm~\ref{alg:dadil}.

\begin{algorithm}[H]
    \caption{\gls{dadil} learning loop.}
    \begin{algorithmic}[1]
        \small
        \REQUIRE $\mathcal{Q} = \{\hat{Q}_{\ell}\}_{\ell=1}^{N}$, number of iterations $N_{iter}$, of atoms $K$, of batches $M$, batch size $n_{b}$, learning rate $\eta$.
        \STATE Initialize $\mathbf{x}_{j}^{(P_{k})}\sim\mathcal{N}(0,\mathbf{I}_{d})$, $\mathbf{a}_{\ell} \sim \mathcal{N}(0,\mathbf{I}_{K})$.
        \FOR{$it=1\cdots,N_{iter}$}
            \FOR{$batch=1,\cdots,M$}
                \FOR{$\ell=1,\cdots,(N_{S}+1)$}
                    \STATE Sample $\{\mathbf{x}_{1}^{(Q_{\ell})},\cdots,\mathbf{x}_{n_{b}}^{(Q_{\ell})}\}$.
                      \IF {$\hat{Q}_{\ell}$ is labeled}
                        \STATE Sample $\{\mathbf{y}_{1}^{(Q_{\ell})},\cdots,\mathbf{y}_{n_{b}}^{(Q_{\ell})}\}$.
                      \ENDIF 
                    \FOR{$k=1,\cdots,K$}
                        \STATE sample $\{(\mathbf{x}_{1}^{(P_{k})}, \mathbf{p}^{(P_{k})}_{1}), \cdots, (\mathbf{x}_{n_{b}}^{(P_{k})}, \mathbf{p}^{(P_{k})}_{n_{b}})\}$,
                        \STATE change variables $\mathbf{y}_{j}^{(P_{k})} = \text{softmax}(\mathbf{p}^{(P_{k})}_{j})$
                    \ENDFOR
                    \STATE calculate $\mathbf{X}^{(B_{\ell})}, \mathbf{Y}^{(B_{\ell})} = \mathcal{B}(\alpha_{\ell};\mathcal{P})$
                \ENDFOR
                \STATE $L = (\nicefrac{1}{N})\sum_{\ell=1}^{N}\mathcal{L}(\hat{Q}_{\ell},\hat{B}_{\ell})$
                \STATE $\mathbf{x}_{j}^{(P_{k})} \leftarrow \mathbf{x}_{j}^{(P_{k})} - \eta \nicefrac{\partial L}{\partial \mathbf{x}^{(P_{k})}_{j}}$
                \STATE $\mathbf{p}_{j}^{(P_{k})} \leftarrow \mathbf{p}_{j}^{(P_{k})} - \eta \nicefrac{\partial L}{\partial \mathbf{p}^{(P_{k})}_{j}}$
                \STATE $\alpha_{\ell} \leftarrow \text{proj}_{\Delta_{K}}(\alpha_{\ell}-\eta \nicefrac{\partial L}{\partial \alpha_{\ell}})$.
            \ENDFOR
        \ENDFOR
        \ENSURE Dictionary $\mathcal{P}^{\star}$ and weights $\mathcal{A}^{\star}$.
    \end{algorithmic}
    \label{alg:dadil}
\end{algorithm}

\begin{wrapfigure}{r}{0.5\linewidth}
    \centering
    \resizebox{\linewidth}{!}{\input{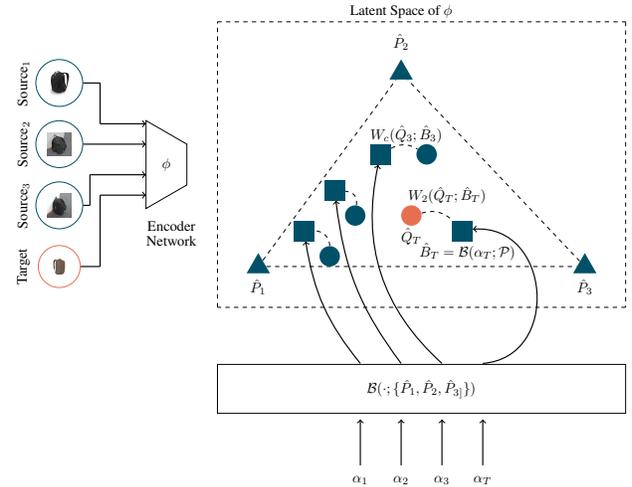}}
    \caption{Conceptual illustration of \gls{dadil}. Each domain is denoted by a blue or orange circle, corresponding to whether it is labeled or not. \gls{dadil} \emph{reconstructs} domains as Wasserstein barycenters, denoted by squares, of atoms, denoted by triangles. The target domain (orange circle) is unlabeled, but we are able to represent it through a labeled distribution through a Wasserstein barycenter.}
    \label{fig:dadil_encoder}
\end{wrapfigure}
\noindent\textbf{Intuition.} We learn how to express each distribution $\hat{Q}_{\ell} \in \mathcal{Q}$ as a barycenter of free distributions $\mathcal{P}=\{\hat{P}_{k}\}_{k=1}^{K}$, parametrized by their support i.e., $(\mathbf{X}^{(P_{k})}, \mathbf{Y}^{(P_{k})})$. In other words, we learn $\mathcal{P}$ s.t. $\mathcal{Q}$ is contained in the \emph{Wasserstein hull} of atoms, $\mathcal{M}(\mathcal{P})$.

\noindent\textbf{Implementation.} We implement algorithms~\ref{alg:wbt_fixed_point_joint} and~\ref{alg:dadil} using Pytorch~\cite{paszke2017automatic} and \gls{pot}~\cite{flamary2021pot}, for automatic differentiation and \gls{ot} details respectively. As previous works~\cite{montesuma2021cvpr, turrisi2022multi}, \gls{dadil} is applied to the latent space of an encoder, pre-trained on source domain data, as shown in figure~\ref{fig:dadil_encoder}.

\noindent\textbf{Computational Complexity.} In algorithm~\ref{alg:dadil}, the complexity of line 13 dominates over other lines. As we discussed in section~\ref{sec:labeled_wbary}, the complexity of calculating $\mathcal{B}(\alpha_{\ell};\mathcal{P})$ depends on the size of distributions support. Since we do computations using mini-batches, this corresponds to $\mathcal{O}(N_{itb}n_{b}^{3}\log n_{b})$. This is repeated for $N_{iter}\times M \times (N_{S} + 1)$, which implies a complexity of $\mathcal{O}(N_{iter}MN_{S}N_{itb}n_{b}^{3} \log n_{b})$.



\noindent\textbf{Multi-Source Domain Adaptation.} We recast the hypothesis in eq.~\ref{eq:hypothesis_courty} for \gls{msda}. We assume the existence of $K > 1$ unknown distributions, $P_{1},\cdots,P_{K}$ for which $Q_{\ell}$ can be approximated as their interpolation in Wasserstein space, i.e. $Q_{\ell}=T_{\sharp}B_{\ell}$, and $Q_{\ell}(Y|X) = B_{\ell}(Y|T(X))$, for $B_{\ell} = \mathcal{B}(\alpha_{\ell};\mathcal{P})$ and a possibly non-linear transformation $T$. If $W_{c}(Q_{\ell}, B_{\ell}) \approx 0$ we can assume $T(\mathbf{x}) = \mathbf{x}$.


\begin{figure}[ht]
    \centering
    \includegraphics[width=0.3\linewidth]{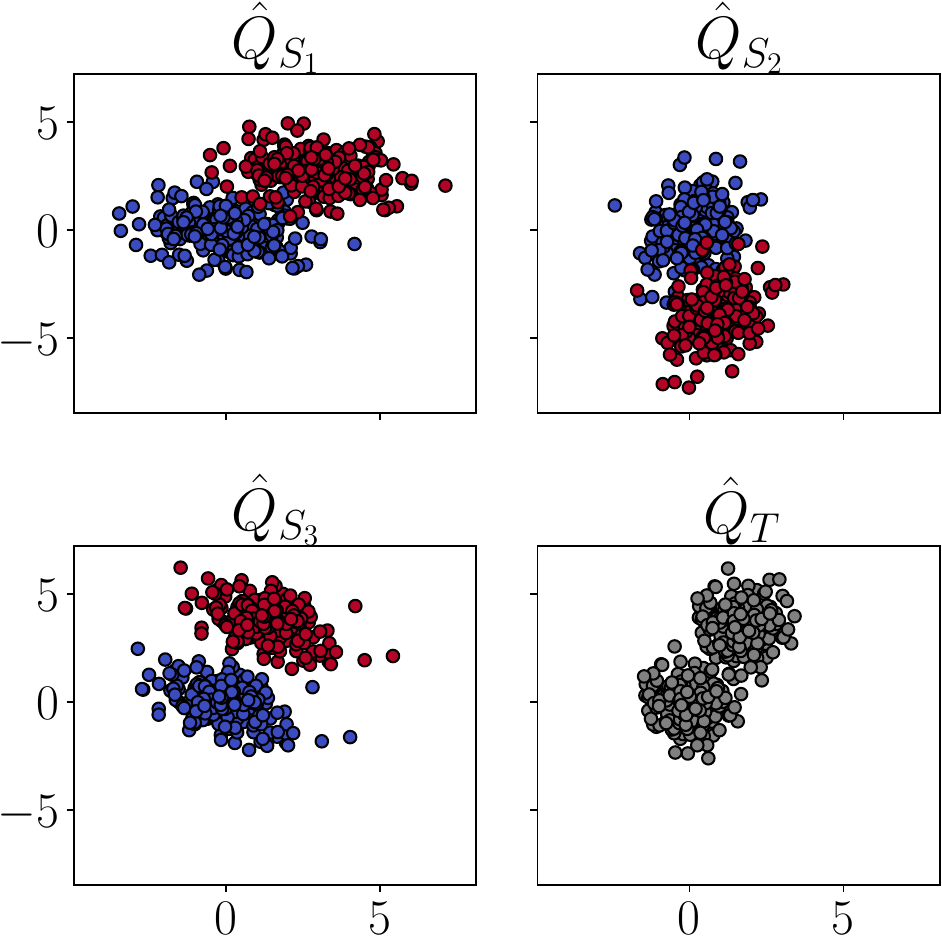}
    \hfill
    \includegraphics[width=0.3\linewidth]{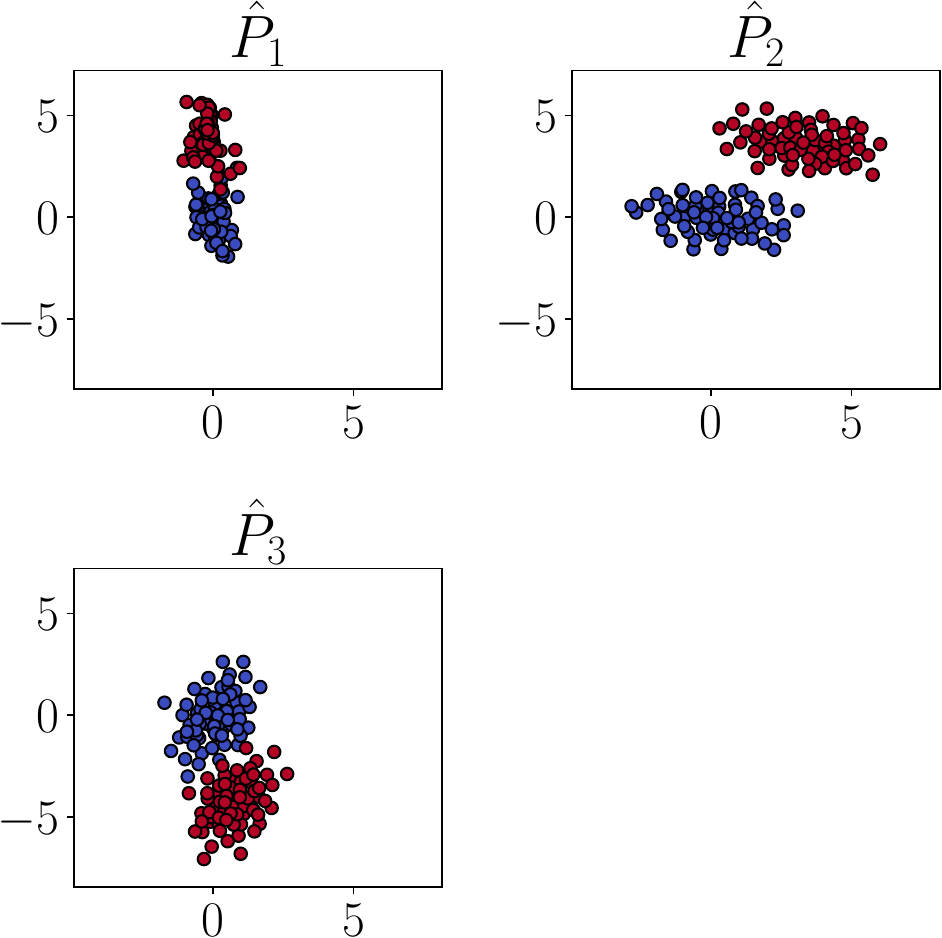} 
    \hfill
    \includegraphics[width=0.3\linewidth]{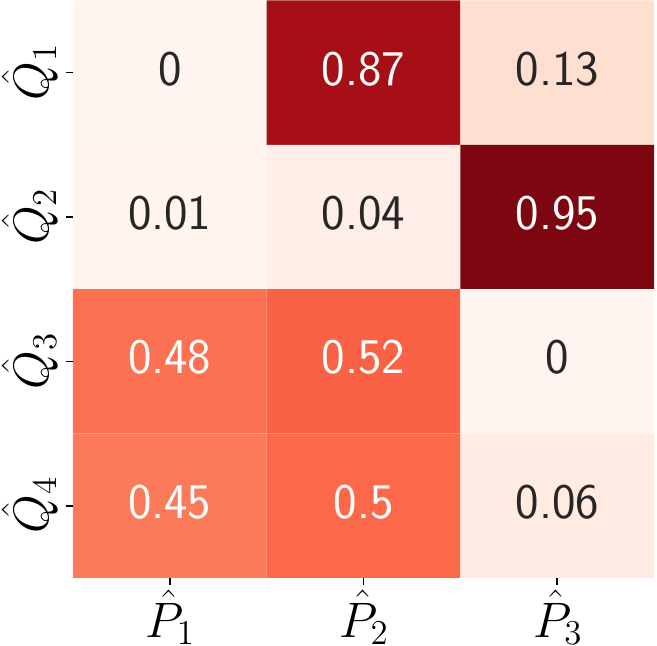}
    \caption{From left to right: set of datasets $\mathcal{Q} = \{\hat{Q}_{S_{1}},\hat{Q}_{S_{2}},\hat{Q}_{S_{3}},\hat{Q}_{T}\}$, where $\hat{Q}_{T}$ is the unlabeled target domain; atoms $\mathcal{P}=\{\hat{P}_{1}, \hat{P}_{2}, \hat{P}_{3}\}$; barycentric weights $\mathcal{A}$.}
    \label{fig:overview_dadil}
\end{figure}

\begin{figure}[ht]
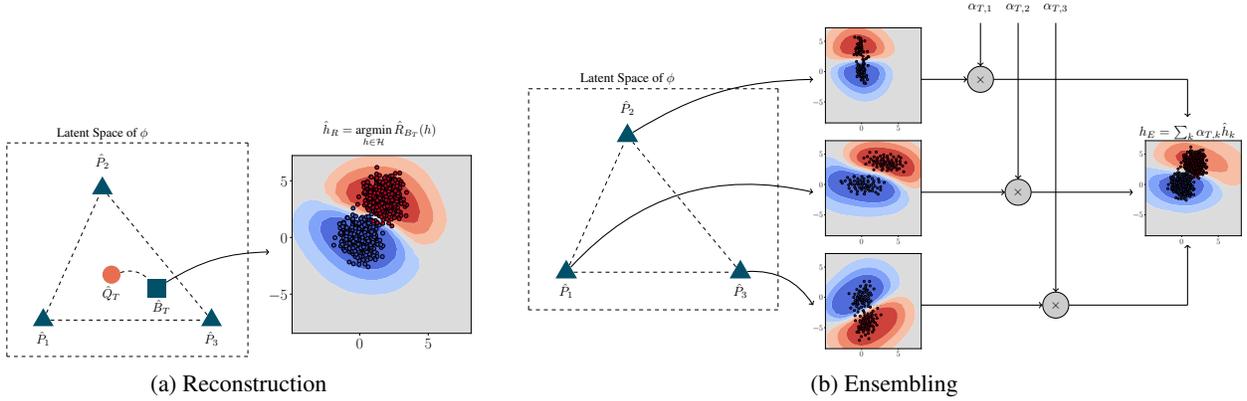

    \centering
    \begin{subfigure}[b]{0.38\linewidth}
        \centering
        \resizebox{\linewidth}{!}{\input{dadil_r.tikz}}
        \caption{Reconstruction}
        \label{fig:dadil_r}
    \end{subfigure}\hfill
    \begin{subfigure}[b]{0.58\linewidth}
        \centering
        \resizebox{\linewidth}{!}{\input{dadil_e.tikz}}
        \caption{Ensembling}
        \label{fig:dadil_e}
    \end{subfigure}
    \caption{Conceptual illustration of the 2 methods, based on \gls{dadil}, for \gls{msda}.}
    \label{fig:dadil_msda}
\end{figure}

We start by learning $(\mathcal{P},\mathcal{A})$, as illustrated in figure~\ref{fig:overview_dadil}. Then, we propose 2 ways of using our dictionary for \gls{msda}. Our first strategy, called \gls{dadil}-R, consists on computing $\hat{B}_{T} = \mathcal{B}(\alpha_{T};\mathcal{P})$, i.e., the distribution in $\mathcal{M}(\mathcal{P})$ closest to $\hat{Q}_{T}$. Since each $\hat{P}_{k}$ has a labeled support, algorithm~\ref{alg:wbt_fixed_point_joint} yields matrices $\mathbf{X}^{(B_{T})}$ and $\mathbf{Y}^{(B_{T})}$ corresponding to the support of $\hat{B}_{T}$. Then,
\begin{align*}
    \hat{h}_{R} = \argmin{h\in\mathcal{H}}\hat{\mathcal{R}}_{B_{T}}(h) = \dfrac{1}{n}\sum_{i=1}^{n}\mathcal{L}(h(\mathbf{x}_{i}^{(B_{T})}),y_{i}^{(B_{T})})
\end{align*}
We theoretically justify it using Theorem 2 of~\cite{redko2017theoretical},
\begin{theorem}{(Due to~\cite{redko2017theoretical})}\label{thm:redko_dadil_r}
Let $\mathbf{X}^{(P)} \in \mathbb{R}^{n_{P} \times d}$ and $\mathbf{X}^{(Q)} \in \mathbb{R}^{n_{Q}\times d}$ be i.i.d. samples from $P$ and $Q$. Then, for any $d' > d$ and $\xi' < \sqrt{2}$ there exists some constant $n_{0}$ depending on $d'$ s.t. for $\delta \in (0, 1)$ and $\text{min}(n_{P},n_{Q}) \geq n_{0}\text{max}(\delta^{-(d+2)},1)$ with probability at least $1 - \delta$ for all $h$,
\begin{align*}
    \mathcal{R}_{Q}(h) &\leq \mathcal{R}_{P}(h) + W_{2}(\hat{P},\hat{Q}) + \zeta + \lambda,
\end{align*}
where,
\begin{align*}
    \zeta &= \sqrt{2\nicefrac{(\log\nicefrac{1}{\delta})}{\xi'}}\biggr(\sqrt{\nicefrac{1}{n_{P}}}+\sqrt{\nicefrac{1}{n_{Q}}}\biggr),\\
    \lambda &= \text{min}_{h\in\mathcal{H}}\mathcal{R}_{Q}(h) + \mathcal{R}_{P}(h).
\end{align*}
\end{theorem}
Additional discussion on this result is provided in our appendix. We apply this result for the residual shift $W_{2}(\hat{Q}_{T},\hat{B}_{T})$,
\begin{align}
    \mathcal{R}_{Q_{T}}(h) \leq \mathcal{R}_{B_{T}}(h) + W_{2}(\hat{Q}_{T},\hat{B}_{T}) + \zeta+\lambda.\label{eq:bound_redko_otda}
\end{align}
As discussed in~\cite{redko2017theoretical}, 3 factors play a role in the success of \gls{da}, namely, $W_{2}(\hat{P},\hat{Q})$, $\mathcal{R}_{B_{T}}(h)$, and $\lambda$. The first term is the reconstruction error, and is directly minimized in algorithm~\ref{alg:dadil}. The second term is the risk of $h$ in $B_{T}$, which is minimized when learning the classifier $\hat{h}_{R} = \text{argmin }\hat{\mathcal{R}}_{B_{T}}(h)$. This term depends on the separability of classes in $\hat{B}_{T}$, which is enforced by considering labels in the ground-cost (eqn.~\ref{eq:supervised_ground_cost}). The last term is the joint risk $\lambda$ of a classifier learned with data from $Q_{T}$ and $B_{T}$. This term is difficult to bound, as no labels in $\hat{Q}_{T}$ are available, but, under the hypothesis $Q_{T}(Y|X)=B_{T}(Y|T(X))$, this term is low. This was similarly assumed by~\cite{courty2016optimal,redko2017theoretical}. \gls{dadil}-R is illustrated in figure~\ref{fig:dadil_r}.

Our second strategy, called \gls{dadil}-E, is based on ensembling. Since each of our atoms is labeled, i.e., each $\mathbf{x}_{i}^{(P_{k})}$ has an associated $\mathbf{y}_{i}^{(P_{k})}$, we may learn a set of $K$ classifiers, $\hat{h}_{k} = \text{argmin}_{h\in\mathcal{H}}\hat{\mathcal{R}}_{P_{k}}(h)$, one for each atom. Naturally, one may use $\alpha \in \Delta_{K}$ for weighting predictions of atom classifiers. We weight the $\hat{h}_{k}$'s using $\alpha_{T}$, which is theoretically justified in theorem~\ref{thm:dadil_guarantee}, 
\begin{align*}
    \hat{h}_{E}(\mathbf{x}^{(Q_{T})}_{j})=\sum_{k=1}^{K}\alpha_{T,k}\hat{h}_{k}(\mathbf{x}_{j}^{(Q_{T})}),
\end{align*}

\begin{theorem}\label{thm:dadil_guarantee}
Let $\{\mathbf{X}^{(P_{k})}\}_{k=1}^{K}$, $\mathbf{X}^{(P_{k})} \in \mathbb{R}^{n_{k} \times d}$ and $\mathbf{X}^{(Q_{T})} \in \mathbb{R}^{n_{T} \times d}$ be i.i.d. samples from $P_{k}$ and $Q_{T}$. Let $\hat{h}_{k}$ be the minimizer of $\mathcal{R}_{P_{k}}$ and $\mathcal{R}_{\alpha}(h) = \sum_{k=1}^{K}\alpha_{k}\mathcal{R}_{P_{k}}(h)$. Under the same conditions of theorem~\ref{thm:redko_dadil_r}, and for $\delta \in (0, 1)$, with probability at least $1- \delta$, the following holds,
\begin{align*}
    \mathcal{R}_{Q_{T}}(\hat{h}_{\alpha}) &\leq \mathcal{R}_{\alpha}(\hat{h}_{\alpha}) + \text{W}_{2}(\mathcal{B}(\alpha;\mathcal{P}), \hat{Q}_{T}) + \gamma + \lambda + \zeta,\\
    \gamma &= \sum_{k=1}^{K}\alpha_{k}\text{W}_{2}(\hat{P}_{k}, \mathcal{B}(\alpha;\mathcal{P})),\\
    \zeta &= \sum_{k=1}^{K}\alpha_{k}\sqrt{\nicefrac{2\log\nicefrac{1}{\delta}}{\xi'}}\biggr(\sqrt{\nicefrac{1}{n_{k}}}+\sqrt{\nicefrac{1}{n_{T}}}\biggr),\\
    \lambda &= \sum_{k=1}^{K}\alpha_{k}\biggr(\text{\text{min}}_{h\in\mathcal{H}}\mathcal{R}_{P_{k}}(h)+\mathcal{R}_{Q_{T}}(h)\biggr).
\end{align*}
\end{theorem}
We provide the proof of this result and additional discussion in our appendix. This bound depends on different terms. First, $\gamma$ is, for a given $\alpha$, minimal, as $\mathcal{B}(\alpha;\mathcal{P})$ is the minimizer of $\hat{B} \mapsto \sum_{k}\alpha_{k}W_{2}(\hat{P}_{k},\hat{B})$. $\lambda$ corresponds to the complexity of domain adaptation, and in general cannot be directly controlled due the unavailability of labels in $\hat{Q}_{T}$. Finally, $\xi$ corresponds to the sample complexity of estimating $W_{2}(P_{k},Q_{T})$ via finite samples. Note that $\alpha_{T}$ minimizes the terms in the r.h.s., as, by design, it minimizes the term $\alpha \mapsto W_{2}(\mathcal{B}(\alpha;\mathcal{P}),\hat{Q}_{T})$. \gls{dadil}-E is illustrated in figure~\ref{fig:dadil_e}.

\section{Experiments}\label{sec:experiments}

\subsection{Multi-Source Domain Adaptation}

\textbf{Experimental Setup.} All experiments were run on a computer with a Ubuntu 22.04 OS, a 12th Gen Intel(R) Core\textsuperscript{TM} i9-12900H CPU with 64 GB of RAM, and with a NVIDIA RTX A100 GPU with 4GB of VRAM. We explore the following hyper-parameters,
\begin{itemize}
    \item Number of samples $n$ is searched among $\{50,100, 200\} \times n_{c}$.
    \item Number of atoms $K$ is searched among $\{3, 4, \cdots, 8\}$.
    \item Batch size $n_{b}$ is searched among $\{5, 10, 20\} \times n_{c}$. We further sample balanced batches from the sources.
\end{itemize}
The complexity of our model is controlled through $n$ and $K$. We provide further analysis on the robustness w.r.t. hyper-parameter choice, as well as the full set of chosen hyper-parameters in our appendix. For other algorithms from the \gls{sota}, we use the best hyper-parameter settings reported by their respective authors.

\noindent\textbf{Caltech-Office 10} is a benchmark consisting on the intersection of the Caltech 256 dataset of~\cite{griffin2007caltech} and the Office 31 dataset of~\cite{saenko2010adapting}. It has 4 domains: Amazon (A), dSLR (D), Webcam (W) and Caltech (C). In this benchmark we compare \gls{dadil} with other shallow \gls{da} algorithms, such as: (i) \gls{sa} of \cite{fernando2013unsupervised}; (ii) \gls{tca} of \cite{pan2010domain}; (iii) \gls{otda} of \cite{courty2016optimal}; (iv) \gls{wbt} of \cite{montesuma2021cvpr,montesuma2021icassp}; (v) \gls{wjdot} of \cite{turrisi2022multi}. (i) and (ii) are standard algorithms in \gls{da}, (iii) is the single-source \gls{ot} baseline, and (iv, v) are the \gls{sota} for shallow \gls{msda}. The baseline corresponds to training a single-layer Perceptron with the concatenation of source domain data.

\begin{table}
    \caption{Classification accuracy (in \%) of \gls{da} methods. Each column represents a target domain for which we report mean $\pm$ standard deviation over $5$ folds. $^{*}$ and $^{\dagger}$ denote results from \protect\cite{montesuma2021cvpr} and \protect\cite{turrisi2022multi}.}
    \centering
    \begin{tabular}{lccccc}
        \toprule
        Method & $A$ & $D$ & $W$ & $C$ & Avg\\
        \midrule
        Baseline & 90.55 $\pm$ 1.36 & 96.83 $\pm$ 1.33 & 88.36 $\pm$ 1.33 & 82.95 $\pm$ 1.26 & 89.67\\
        \midrule
        SA & 88.61 $\pm$ 1.72 & 92.08 $\pm$ 3.82 & 79.33 $\pm$ 3.67 & 73.00 $\pm$ 2.31 & 83.26 \\
        TCA$^{\star}$ & 86.83 $\pm$ 4.71 & 89.32 $\pm$ 1.33 & 97.51 $\pm$ 1.18 & 80.79 $\pm$ 2.65 & 88.61 \\
        OTDA & 88.26 $\pm$ 1.36 & 90.41 $\pm$ 3.86 & 88.09 $\pm$ 3.80 & 83.02 $\pm$ 1.67 & 87.44 \\
        \midrule
        WJDOT$^{\dagger}$ & \textbf{94.23} $\pm$ \textbf{0.90} & \textbf{100.00} $\pm$ \textbf{0.00} & 89.33 $\pm$ 2.91 & 85.93 $\pm$ 2.07 & 92.37 \\
        WBT$_{reg}^{\star}$ & 92.74 $\pm$ 0.45 & 95.87 $\pm$ 1.43 & 96.57 $\pm$ 1.76 & 85.01 $\pm$ 0.84 & 92.55\\
        DaDiL-R & 94.06 $\pm$ 1.82 & \underline{98.75} $\pm$ \underline{1.71} & \underline{98.98} $\pm$ \underline{1.51} & \underline{88.97} $\pm$ \underline{1.06} & \underline{95.19}\\
        DaDiL-E & \underline{94.16} $\pm$ \underline{1.58} & \textbf{100.00} $\pm$ \textbf{0.00} & \textbf{99.32} $\pm$ \textbf{0.93} & \textbf{89.15} $\pm$ \textbf{1.68} & \textbf{95.66}\\
        \bottomrule
    \end{tabular}
    \label{tab:results_caltech_office}
\end{table}

Our results is presented in table~\ref{tab:results_caltech_office}. \gls{dadil} improve over previous \gls{ot}-based \gls{msda} baselines, i.e. \gls{wjdot} and \gls{wbt}, being especially better on the Webcam and Caltech domains. Overall, we improve previous \gls{sota} by 3.15 in terms of average \gls{da} performance.

\noindent\textbf{Ablation Study.} We investigate the effectiveness of \gls{dil} in comparison with other barycenter-based approaches. As follows, we compare the performance on the Caltech-Office 10 benchmark of 4 methods: (i) \gls{wb}; (ii) \gls{wbt}; (iii) \gls{wbr}-R and \gls{wbr}-E, which can be understood as the adaptation of the framework of~\cite{bonneel2016wasserstein} for point clouds. The R and E methods are analogous to \gls{dadil} when the atoms are initialized and fixed as the source domains. We provide further details of this adaptation in our appendix.

\begin{table}
    \caption{Classification accuracy (in \%) of \gls{da} methods. $\mathcal{P}$ and $\mathcal{A}$ indicate learning atom distributions and barycentric coefficients respectively. $T$ indicates an additional transport step towards $Q_{T}$.}
    \centering
    \begin{tabular}{lcccccccc}
         \toprule
         Method & $\mathcal{P}$ & $\mathcal{A}$ & $T$ & A & D & W & C & Avg. \\
         \midrule
         WB &&&& 88.54 $\pm$ 1.16 & 90.62 $\pm$ 8.38 & 93.89 $\pm$ 3.30 & 83.73 $\pm$ 1.49 & 89.19 \\
         WBT$_{reg}$ &&& \faCheck & 92.74 $\pm$ 0.45 & 95.87 $\pm$ 1.43 & 96.57 $\pm$ 1.76 & 85.01 $\pm$ 0.84 & 92.55\\
         WBR-R &  & \faCheck & & 91.35 $\pm$ 1.19 & 91.87 $\pm$ 9.47 & 81.69 $\pm$ 3.26 & 86.31 $\pm$ 1.73 & 86.09\\
         WBR-E &  & \faCheck & & 91.97 $\pm$ 2.40 & 91.87 $\pm$ 2.79 & 83.73 $\pm$ 2.57 & 86.13 $\pm$ 1.84 & 88.42\\
        DaDiL-R & \faCheck & \faCheck & &
 \underline{94.06} $\pm$ \underline{1.82} & \underline{98.75} $\pm$ \underline{1.71} & \underline{98.98} $\pm$ \underline{1.51} & \underline{88.97} $\pm$ \underline{1.06} & \underline{95.19}\\
        DaDiL-E & \faCheck & \faCheck & & \textbf{94.16} $\pm$ \textbf{1.58} & \textbf{100.00} $\pm$ \textbf{0.00} & \textbf{99.32} $\pm$ \textbf{0.93} & \textbf{89.15} $\pm$ \textbf{1.68} & \textbf{95.66}\\
         \bottomrule
    \end{tabular}
    \label{tab:ablation_caltech_office}
\end{table}

We report our findings in table~\ref{tab:ablation_caltech_office}. Overall, \gls{wb} and \gls{wbr} have sub-optimal performance. On the one hand, this implies that $\hat{Q}_{T} \not\in \mathcal{M}(\mathcal{Q}_{S})$. On the other hand, this implies that \gls{dil} is key for \gls{msda}. Indeed, since $\hat{P}_{k} \in \mathcal{P}$ are free, \gls{dadil} learns $\mathcal{P}$ s.t. $\hat{Q}_{T} \in \mathcal{M}(\mathcal{P})$. \gls{wbt}$_{reg}$ compensates this fact by transporting the $\hat{B}$ towards $\hat{Q}_{T}$, thus minimizing the \emph{residual shift} $W_{2}(\hat{B},\hat{P}_{T})$.

\noindent\textbf{Refurbished Office 31.} In this experiment, we use the Office 31 benchmark of~\cite{saenko2010adapting}, with the improvements proposed by~\cite{ringwald2021adaptiope}. This benchmark has 3 domains: Amazon (A), dSLR (D) and Webcam (W). Our goal is to establish a comparison with deep \gls{da} methods. As follows, we consider: (i) \gls{dann} of~\cite{ganin2016domain}, (ii) \gls{wdgrl} of~\cite{shen2018wasserstein}, (iii) Deep-\gls{jdot} of~\cite{damodaran2018deepjdot}, (iv) \gls{m3sda} of~\cite{peng2019moment}, (v) \gls{wjdot} and (vi) \gls{wbt}. While (i) - (iii) are single source baselines, (iv) is a standard method for \gls{msda}. We use a ResNet-50~\cite{he2016deep} as backbone.

\begin{table}
    \caption{Classification accuracy (in \%) of \gls{da} methods on the Refurbished Office 31 benchmark. Each column represents a target domain for which we report mean $\pm$ standard deviation over $5$ folds.}
    \centering
    \begin{tabular}{lcccc}
        \toprule
        Method & $A$ & $D$ & $W$ & Avg\\
        \midrule
        Baseline & 70.57 & 97.00 & 95.47 & 87.68 \\
        \midrule
        DANN & 78.19 & 97.00 & 93.08 & 89.42\\
        WDGRL & 76.06 & 97.00 & 93.71 & 88.92\\
        DeepJDOT & 80.85 & 94.00 & 93.38 & 89.61\\
        \midrule
        M3SDA & 64.89 & \textbf{98.00} & \underline{96.85} & 86.58\\
        WBT$_{reg}$ & 77.48 & 96.00 & 95.59 & 89.69\\
        WJDOT & 70.21 & 97.00 & 94.96 & 87.39\\
        DaDiL-R & \textbf{85.46} & 93.00 & \textbf{97.48} & \textbf{91.98}\\
        DaDiL-E & \underline{83.51} & 94.00 & 94.34 & \underline{90.61}\\
        \bottomrule
    \end{tabular}
    \label{tab:results_office}
\end{table}

A summary of our results is shown in table~\ref{tab:results_office}. Overall, \gls{dadil}-R and E are especially better than previous algorithms in the Amazon domain. As a consequence, in terms of average domain performance, \gls{dadil}-R and E improve over the second-best method (WBT$_{reg}$) by a margin of 2.29\% and 1.37\% respectively.

\noindent\textbf{CWRU.} In this benchmark, we explore \gls{dadil} for cross-domain fault diagnosis. The goal is to classify which type of fault has occurred, based on sensor readings. Hence, we extract $2048$ Fourier coefficients from a sub-set of $4096$ time-steps extracted from the raw signals (see~\cite{zhang2018intelligent}, or our appendix for more details). As feature extractor, we use a 3-layer fully connected encoder. We compare 3 single, and 5 multi-source \gls{da} algorithms to \gls{dadil}, namely, \gls{dann}, \gls{otda}, \gls{tca}, \gls{m3sda}, LTC-MSDA of~\cite{wang2020learning}, JCPOT of~\cite{redko2019advances}, \gls{wbt}$_{reg}$ and \gls{wjdot}.

\begin{table}
    \caption{Classification accuracy (in \%) of \gls{da} methods on the CWRU benchmark. Each column represents a target domain for which we report mean $\pm$ standard deviation over $5$ folds.}
    \centering
    \begin{tabular}{lcccc}
        \toprule
        Method & $1772$rpm & $1750$rpm & $1730$rpm & Avg\\
        \midrule
        Baseline & 70.90 $\pm$ 0.40 & 79.76 $\pm$ 0.11 & 72.26 $\pm$ 0.23 & 74.31 \\
        \midrule
        DANN & 67.96 $\pm$ 8.52 & 64.38 $\pm$ 5.03 & 57.75 $\pm$ 17.06 & 63.37\\
        OTDA & 70.48 $\pm$ 2.25 & 79.61 $\pm$ 0.25 & 74.98 $\pm$ 1.26 & 75.02\\
        TCA & 87.17 $\pm$ 4.25 & 84.11 $\pm$ 4.77 & 92.74 $\pm$ 4.12 & 88.01\\
        \midrule
        M3SDA & 56.86 $\pm$ 7.31 & 69.81 $\pm$ 0.36 & 61.06 $\pm$ 6.35 & 62.57\\
        WJDOT & 65.01 $\pm$ 0.27 & 69.81 $\pm$ 0.07 & 57.40 $\pm$ 1.18 & 64.07\\
        M3SDA$_{\beta}$ & 60.15 $\pm$ 8.38 & 70.00 $\pm$ 0.00 & 64.00 $\pm$ 5.47 & 64.72\\
        LTC-MSDA & 82.21 $\pm$ 8.03 & 75.33 $\pm$ 5.91 & 81.04 $\pm$ 5.45 & 79.52\\
        JCPOT & 77.48 $\pm$ 0.86 & \underline{96.00} $\pm$ \underline{0.10} & 95.59 $\pm$ 0.56 & 91.74\\
        WBT$_{reg}$ & \underline{99.28} $\pm$ \underline{0.18} & 79.91 $\pm$ 0.04 & 97.71 $\pm$ 0.76 & 92.30\\
        DaDiL-R & \textbf{99.86} $\pm$ \textbf{0.21} & \textbf{99.85} $\pm$ \textbf{0.08} & \textbf{100.00} $\pm$ \textbf{0.00}  & \textbf{99.90}\\
        DaDiL-E & 93.71 $\pm$ 6.50 & 83.63 $\pm$ 4.98 & \underline{99.97} $\pm$ \underline{0.05}  & \underline{92.33}\\
        \bottomrule
    \end{tabular}
    \label{tab:results_crwu}
\end{table}

We present a summary of our results in table~\ref{tab:results_crwu}. Overall, \gls{wbt}$_{reg}$ and \gls{dadil} are the best performing methods, demonstrating the power of Wasserstein barycenters for \gls{da}. Our method outperforms \gls{wbt}$_{reg}$ by 7.71\%, in terms of average domain performance. Furthermore, our methods surpass other deep learning baselines, such as \gls{m3sda}~\cite{peng2019moment} and LTC-MSDA~\cite{wang2020learning}, by a margin of 19.90\%.

\subsection{Domain Adaptation using Atom Interpolations}\label{sec:dadil_interpolations}

\begin{figure}[ht]
    \centering
    \includegraphics[width=0.85\linewidth]{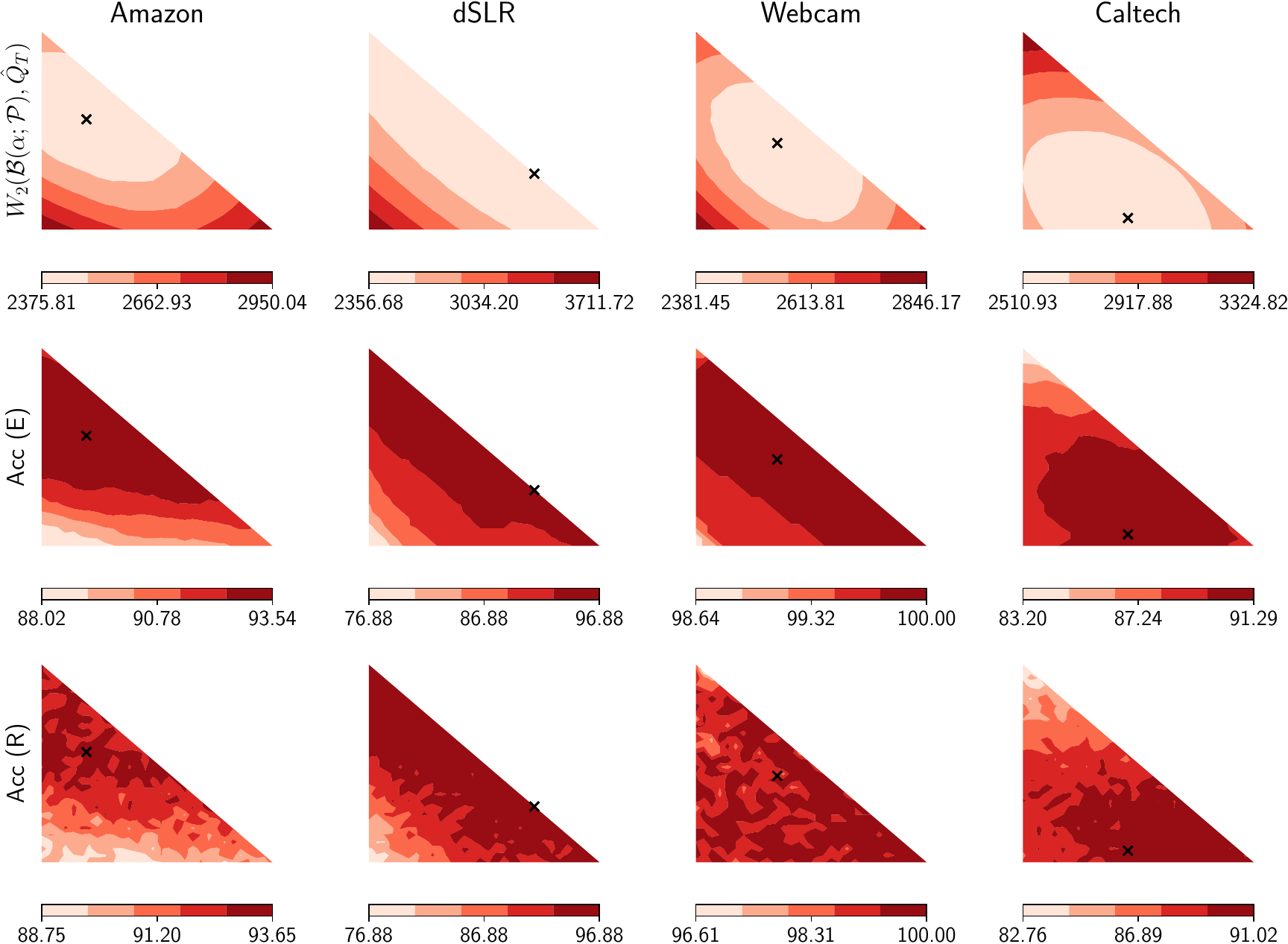}
    \caption{Analysis of \gls{da} on Caltech-Office with interpolations of dictionary atoms. The black cross represents the $\alpha$ found by \gls{dadil}.}
    \label{fig:dadil_interpolation_space}
\end{figure}

Besides performing \gls{msda} with optimal barycentric coordinates $\alpha_{T} \in \Delta_{K}$, in this section we explore the question \emph{how well do $\alpha \in \Delta_{K}$ perform?} We explore these questions in terms of Wasserstein distance $W_{2}(\mathcal{B}(\alpha;\mathcal{P}), \hat{Q}_{T})$, and classification accuracy of using $\alpha$ in \gls{dadil}-R and E, as shown in figure~\ref{fig:dadil_interpolation_space}.

In figure~\ref{fig:dadil_interpolation_space} we construct an uniform grid over $\Delta_{3}$. For each $\alpha$ in such grid, we reconstruct $\mathcal{B}(\alpha;\mathcal{P})$, then we evaluate: (i) the \emph{reconstruction loss} $W_{2}(\mathcal{B}(\alpha;\mathcal{P}),\hat{Q}_{T})$; (ii) the classification accuracy of \gls{dadil}-E, with $\alpha$, on $\hat{Q}_{T}$; (iii) the classification accuracy of \gls{dadil}-R with $\alpha$, on $\hat{Q}_{T}$. These correspond to the 3 rows in figure~\ref{fig:dadil_interpolation_space}. As shown, the weights found by \gls{dadil} are optimal w.r.t. other choices $\alpha \in \Delta_{3}$. Nonetheless, a wide region of the simplex yield \emph{equally good} reconstructions, either w.r.t. reconstruction loss, or w.r.t. \gls{da} performance. We conclude that \gls{dadil} is able to learn distribution whose interpolations generalize well to the target domain.

\begin{figure}[ht]
    \centering
    \includegraphics[width=0.8\linewidth]{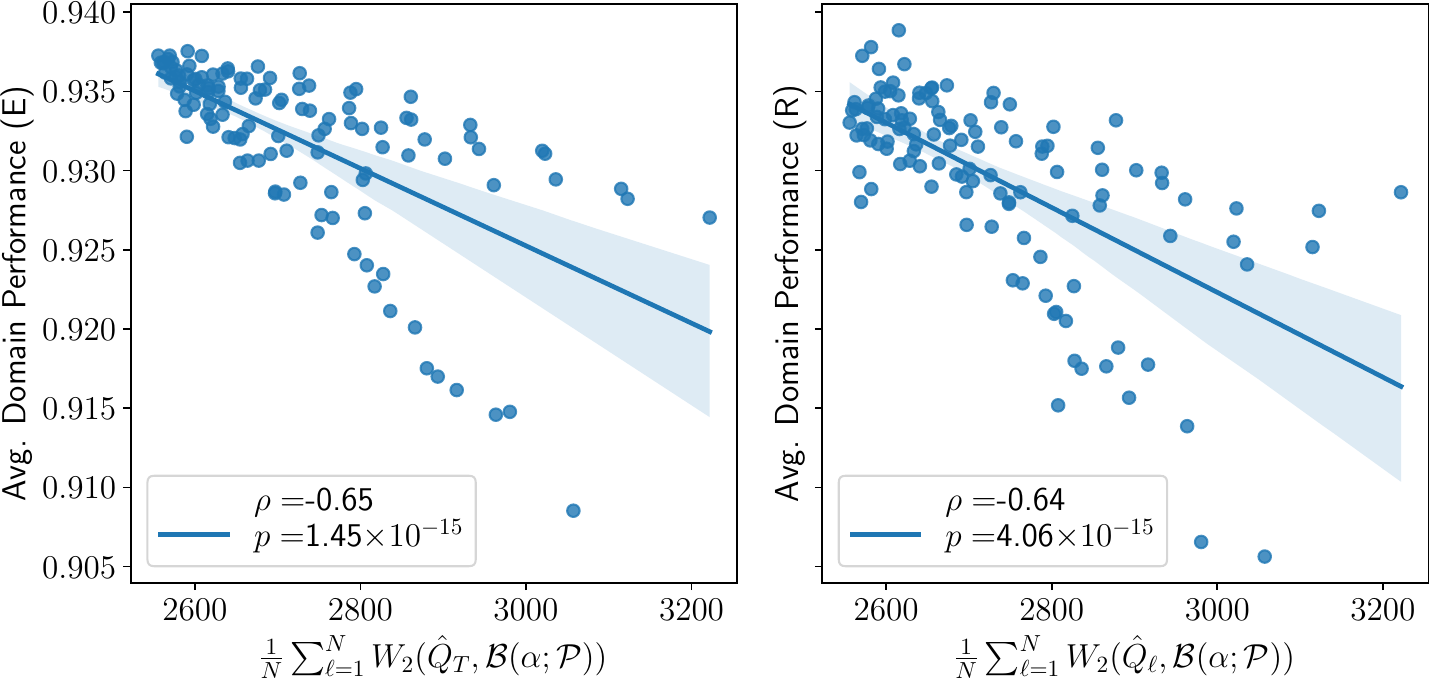}
    \caption{Correlation between \gls{dil} loss and the performance of \gls{dadil}-R and E.}
    \label{fig:corr_acc_loss}
\end{figure}

Furthermore, in figure~\ref{fig:corr_acc_loss} we analyze the correlation between \gls{da} performance and reconstruction loss, for $\alpha \in \Delta_{3}$. Our analysis shows that these 2 terms are negatively correlated, for both \gls{dadil}-R and E. Indeed, based on our theoretical analysis (theorems~\ref{thm:redko_dadil_r} and~\ref{thm:dadil_guarantee}), classification risk is bounded by the reconstruction loss. Since \gls{da} performance is inversely proportional to the classifier risk in a given domain, our analysis agrees with both theorems.

\begin{figure}[ht]
    \centering
\includegraphics[width=\linewidth]{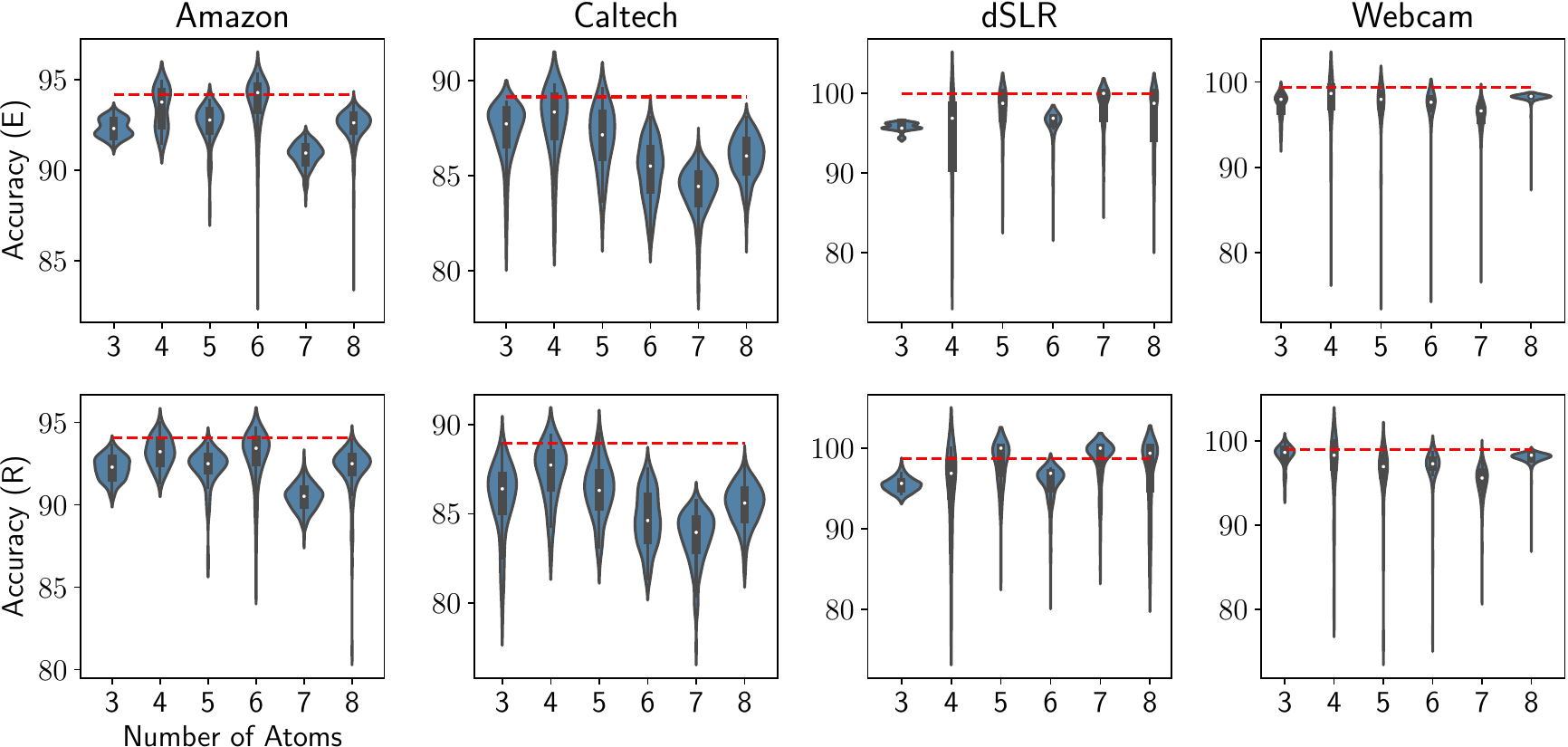}
    \caption{Performance analysis of latent space interpolations on the Caltech-Office 10 benchmark. The red dotted line corresponds to the results reported in Table~\ref{tab:results_caltech_office} for \gls{dadil}.}
    \label{fig:interpolation_avg}
\end{figure}

Finally, we analyze the performance of \gls{dadil}-R and E for $\alpha$ taken uniformly from $\Delta_{K}$, for $K\in\{3,\cdots,8\}$. We report our findings in figure~\ref{fig:interpolation_avg}, and compare the performance w.r.t. \gls{dadil} performance in table~\ref{tab:results_caltech_office}, for $\alpha := \alpha_{T}$. As shown in Figure~\ref{fig:interpolation_avg} $\alpha_{T}$ is above average for most domains and number of atoms $K$.

Overall, figures~\ref{fig:dadil_interpolation_space},~\ref{fig:corr_acc_loss} and~\ref{fig:interpolation_avg} show that \gls{dadil} learns an optimal set of barycentric coordinates for the target domain. Nonetheless, interpolations in the Wasserstein hull $\mathcal{M}(\mathcal{P})$ of atom distributions can be equally interesting for \gls{msda}. These remarks indicate that \gls{dadil} is able to (i) learn common discriminant information about the source domains; (ii) interpolate the distributional shift between the various distributions in $\mathcal{Q} = \{\hat{Q}_{S_{\ell}}\}_{\ell=1}^{N_{S}} \cup \{\hat{Q}_{T}\}$ through the atoms $\mathcal{P}$.

\section{Discussion}\label{sec:discussion}

\noindent\textbf{Benefits of Dictionary Learning.} Our proposed framework allows for the learning of new, \emph{virtual distributions}, which can then be used to reconstruct distributions seen during \gls{dil} by generating new samples. As such, our algorithm is able to improve over past \gls{sota}, and, as shown in section~\ref{sec:dadil_interpolations}, we are able to generate new domains by interpolating the atom distributions in Wasserstein space. Especially, we improve previous \gls{sota} and barycenter-based algorithms by 3.15\% in the Caltech-Office 10 benchmark.

\noindent\textbf{Benefits of Wasserstein Barycenters.} In our experiments, we established a comparison between \gls{dadil}, \gls{wbt}~\cite{montesuma2021cvpr} and \gls{wjdot}~\cite{turrisi2022multi}. The first two methods rely on Wasserstein barycenters for reconstructing the target domain, while \gls{wjdot} aggregates the source domains linearly. Overall we show that Wasserstein barycenters are an important component of \gls{msda}, as they allow to \emph{average probability distributions non-linearly}. On the other hand, the linear average of distributions can be understood as importance weighting on samples. Under the covariate shift hypothesis, re-weighting samples is enough, but under more complicated shifts (i.e. non-linear data transformations), Wasserstein barycenters are more flexible.

\noindent\textbf{Shallow vs. Deep Domain Adaptation.} As remarked by~\cite{turrisi2022multi}, the assumption of having a meaningful feature extractor $\phi$ \emph{before performing \gls{da}} is realistic, as in modern practice pre-trained models are widely available. It is noteworthy that a fine-tuning step with source-domain data may be necessary in order to achieve better performance. In addition, doing so allows for the comparison with deep \gls{da} methods. In this context, we remark that our method improves over previous deep \gls{da} \gls{sota} in the context of the Refurbished Office 31 and CWRU benchmarks. Overall, shallow \gls{da} is computationally simpler than deep \gls{da}, as one needs to learn a smaller set of parameters (i.e., the classification layer).

\section{Conclusion}\label{sec:conclusion}

In this work, we tackle the problem of \gls{msda} through \gls{ot}-based \gls{dil} of probability distributions. We view elements in \gls{dil} as empirical distributions. As such we learn a dictionary that is able to interpolate the distributional shift of distributions in \gls{dil}. We make 2 methodological contributions to \gls{msda}, through methods called \gls{dadil}-R, based on the reconstruction of labeled samples in the target domain, and \gls{dadil}-E, based on ensembling of atom classifiers. Our methods are theoretically grounded on previous theorems from the literature~\cite[Theorem 2]{redko2017theoretical} and a novel result (theorem~\ref{thm:dadil_guarantee}).

Our proposed methods are compared to 11 methods from the \gls{sota} in \gls{msda} in 3 benchmarks, namely, Caltech-Office 10~\cite{saenko2010adapting,griffin2007caltech}, Refurbished Office 31~\cite{saenko2010adapting,ringwald2021adaptiope} and CWRU. We improve previous performance by 3.15\%, 2.29\% and 7.71\% respectively. Moreover, we show that general interpolations in the Wasserstein hull of our learned dictionary can be equally interesting for \gls{msda}.



Our framework opens an interesting line of research, for \emph{learning} empirical distributions, generating synthetic through Wasserstein barycenters and interpolating distributional shift in Wasserstein space. It is flexible so as to accommodate other notions of barycenters of distributions, and loss functions between reconstructions and real datasets. In practical terms, future works will focus on parametric formulations of \gls{dadil}. In theoretical terms, we seek to understand the statistical challenges posed by \gls{dadil}.

\bibliographystyle{unsrt}
\bibliography{references}

\appendix

\section{Introduction}

We divide our supplementary material into 3 sections. In appendix~\ref{appx:proofs} we provide further details about our theoretical results. Especially, appendix~\ref{appx:proof_thm2} we provide a proof for our theorem 2. In appendix~\ref{appx:wbr} we provide a detailed description of our adaptation of the \gls{wbr} algorithm of~\cite{bonneel2016wasserstein}. This algorithm allows us to establish an illustration on \emph{why perform dictionary learning}, as learning atom distributions improves over simply regressing barycentric coordinates of the Wasserstein hull of source distributions $\mathcal{Q}_{S}$. Finally, appendix~\ref{appx:additional_exps} provides further experimentation on the hyper-parameters of our algorithm. We provide in table~\ref{tab:hyperparam_info} the complete setting of hyper-parameters used in our experiments.

\section{Proofs of Theorems}\label{appx:proofs}

\subsection{Preliminaries and Theorem 1}

In the following discussion, we need to extend the definition of risk of a classifier $h$, for pairs $(h, h') \in \mathcal{H}$, as is done in~\cite{redko2017theoretical},
\begin{align*}
    h^{\star} = \argmin{h\in\mathcal{H}}\mathcal{R}_{Q}(h,h')=\expectation{\mathbf{x}\sim Q}[\mathcal{L}(h(\mathbf{x}),h'(\mathbf{x}))].
\end{align*}
This is different to what was presented in section 3.1, but is needed to state the results that follow. Note that, under the assumption that $\mathcal{L}$ suffices the triangle inequality, so does $\mathcal{R}_{Q}$ for $h_{1}, h_{2}, h_{3} \in \mathcal{H}$. We start by bounding risks under $P$ and $Q$ by their Wasserstein distance.

\begin{lemma}(Due to~\cite{redko2017theoretical})\label{lemma:bound_risk}
Let $P$ and $Q$ be two probability distributions over $\mathbb{R}^{d}$. Assume that the cost function $c(\mathbf{x}^{(P)},\mathbf{x}^{(Q)}) = \lVert\phi(\mathbf{x}^{(P)}) - \phi(\mathbf{x}^{(Q)})\rVert_{\mathcal{F}}$, where $\mathcal{F}$ is a reproducing kernel Hilbert space equipped with kernel $\Phi:\mathbb{R}^{d}\times\mathbb{R}^{d}\rightarrow\mathbb{R}$ induced by $\phi:\mathbb{R}^{d}\rightarrow\mathcal{H}_{k}$ and $\Phi(\mathbf{x}^{(P)},\mathbf{x}^{(Q)}) = \langle \phi(\mathbf{x}^{(P)}), \phi(\mathbf{x}^{(Q)}) \rangle_{\mathcal{F}}$. Assume that the kernel $\Phi \in \mathcal{F}$ is square-root integrable w.r.t. both $P$ and $Q$ and $0 \leq \Phi(\mathbf{x}^{(P)},\mathbf{x}^{(Q)}) \leq M$, $\forall \mathbf{x}^{(P)},\mathbf{x}^{(Q)} \in \mathbb{R}^{d}$. Then the following holds,
\begin{align}
    \mathcal{R}_{Q}(h,h') \leq \mathcal{R}_{P}(h, h') + W_{1}(P, Q).\label{eq:bound_risks}
\end{align}
\end{lemma}
\textbf{Note.} Following~\cite{villani2009optimal}, Hölder's inequality implies that $p \leq q \implies W_{p} \leq W_{q}$, where $W_{p}$ corresponds to the Wasserstein distance with ground-cost $c(\mathbf{x}^{(P)},\mathbf{x}^{(Q)}) = \lVert \mathbf{x}^{(P)} - \mathbf{x}^{(Q)}\rVert_{p}^{p}$. As consequence, bound~\ref{eq:bound_risks} is also valid for the Euclidean distance.

The consequence of Lemma~\ref{lemma:bound_risk} relates the risk of a pair $(h, h')$ under distributions $P$ and $Q$ by the Wasserstein distance. This is in line with other theoretical results, such as~\cite{ben2010theory}. Before relating the risks of $h$ with the \emph{empirical Wasserstein distance}, one needs to establish the convergence of the empirical $\hat{P}$ to 
 the true $P$. This is done in the next lemma.


\begin{lemma}\label{lemma:bolley_uniform_convergence}
(Due to~\cite{bolley2007quantitative}) Let $P$ be a probability distribution over $\mathbb{R}^{d}$, so that for some $\alpha > 0$ we have that $\int_{\mathbb{R}^{d}}e^{\alpha||\mathbf{x}||^{2}}dP < \infty$ and $\hat{P}$ be its associated empirical approximation with support $\{\mathbf{x}_{i}^{(P)}\}_{i=1}^{n}$ drawn independently from $P$. Then, for any $d' > d$ and $\xi' < \sqrt{2}$ there is a constant $n_{0}$ depending on $d'$ and some square exponential moment of $P$ such that for any $\epsilon > 0$ and $n \geq n_{0}\text{max}(\epsilon^{-(d'+2)}, 1)$,
\begin{align*}
    \mathbb{P}[W_{1}(\hat{P}, P) > \epsilon] \leq \exp\biggr(-\dfrac{\xi'}{2}n\epsilon^{2}\biggr),
\end{align*}

\noindent where $d'$ and $\xi'$ can be calculated explicitly.

\end{lemma}

Lemma~\ref{lemma:bolley_uniform_convergence} states the conditions for which $\hat{P}$ and $P$ are close in the sense of Wasserstein. This last bound is on the form $\mathbb{P}[quantity>\epsilon] < \delta$, that is, with high probability $quantity \leq \epsilon$. These types of bounds are ubiquitous in the theoretical analysis of learning algorithms. We can express $\epsilon$ explicitly in terms of $\delta$,
\begin{align}
    \epsilon &= \sqrt{\dfrac{2}{n\xi'}\log(\scriptsize\tfrac{1}{\delta})},\label{eq:uniform_convergence}
\end{align}
which will be useful in the following discussion. We recall that, as stated in~\cite[Theorem 2]{redko2017theoretical}, and Theorem~\ref{thm:redko_dadil_r}, under suitable conditions,
\begin{align*}
    \mathcal{R}_{Q}(h) &\leq \mathcal{R}_{P}(h) + W_{2}(\hat{P},\hat{Q}) + \zeta + \lambda,\\
    \zeta &= \sqrt{2\nicefrac{(\log\nicefrac{1}{\delta})}{\xi'}}\biggr(\sqrt{\nicefrac{1}{n_{P}}}+\sqrt{\nicefrac{1}{n_{Q}}}\biggr),\\
    \lambda &= \text{min}_{h\in\mathcal{H}}\mathcal{R}_{Q}(h) + \mathcal{R}_{P}(h),
\end{align*}

These results allowed~\cite{redko2017theoretical} to provide theoretical guarantees for the \gls{otda} framework of~\cite{courty2016optimal}. As discussed in~\cite{redko2017theoretical}, the minimization of $W_{2}(\hat{P},\hat{Q})$ alone does not guarantees \gls{da} success. As illustrated by the authors, during transportation one can minimize the domain distance while mixing classes. This leads to a situation where one has a low $W_{2}$, but a risk $\mathcal{R}_{P}$ high over all $h \in \mathcal{H}$. The same reasoning may be applied to the joint error $\lambda$. A few assumptions need to be done at this point,
\begin{enumerate}
    \item[H1] The distribution $\hat{P}'$ which reduces $W_{2}(\cdot,\hat{Q})$ has distinguishable classes, i.e., $\mathcal{R}_{P'}(h)$ is low.
    \item[H2] The distribution $\hat{P}'$ which reduces $W_{2}(\cdot,\hat{Q})$ has the same class-structure as $\hat{Q}$ (i.e. classes lie on the same side of the boundary). In other words, $\lambda$ is low.
\end{enumerate}
We now recast this theorem and hypothesis in terms of \gls{dadil}-R. First, one has $Q = Q_{T}$, i.e., the target domain distribution. Second, one uses $P = B_{T}$, i.e., the barycentric reconstruction of the target, through our dictionary. Our algorithm regulates $\mathcal{R}_{B_{T}}$ by integrating the labels in the ground-cost. As a consequence, giving mass to samples with different classes is too costly, and the \gls{ot} transport plans are \emph{class sparse}, i.e. $\pi_{ij} > 0 \iff y_{i}^{(P)} = y_{j}^{(Q)}$.

The second hypothesis is difficult to ensure in unsupervised \gls{da}, as one does not have any information about the class structure of $\hat{Q}$. As~\cite{courty2016optimal} and~\cite{redko2017theoretical}, we assume a degree of regularity in the distributional shift of distributions in $\mathcal{Q}$. This allows us to predict classes in the target, through label propagation.

\subsection{Proof of Theorem 2}\label{appx:proof_thm2}

Before proving our results, we define the following risks,
\begin{align*}
    \mathcal{R}_{Q_{T}}(h) &= \expectation{\mathbf{x}\sim Q_{T}}[\mathcal{L}(h(\mathbf{x}), h_{Q_{T},0}(\mathbf{x}))],\\
    \mathcal{R}_{P_{k}}(h) &= \expectation{\mathbf{x}\sim P_{k}}[\mathcal{L}(h(\mathbf{x}), h_{P_{k},0}(\mathbf{x}))],\\
    \mathcal{R}_{\alpha}(h) &= \sum_{k=1}^{K}\alpha_{k}\mathcal{R}_{P_{k}}(h).
\end{align*}
which are the risk under the target, under each of the atom distributions, and the combined risk weighted by $\alpha$, respectively. $h_{Q_{T},0}$ and $h_{P_{k},0}$ are the ground-truth labeling functions of the target distribution, and atom $k$. Likewise, we define the following classifiers,
\begin{align*}
    h_{T,k}^{\star} &= \text{argmin}_{h\in\mathcal{H}}\mathcal{R}_{Q_{T}}(h)+\mathcal{R}_{P_{k}}(h),\\
    \hat{h}_{k} &= \text{argmin}_{h\in\mathcal{H}}\mathcal{R}_{P_{k}}(h),\\
    \hat{h}_{\alpha}(\mathbf{x}) &= \sum_{k}\alpha_{k}\hat{h}_{k}(\mathbf{x}).
\end{align*}
Our proof relies on the triangle inequality for the risk. With our previous definitions,
\begin{equation*}
    \mathcal{R}_{Q_{T}}(\hat{h}_{\alpha}) \leq \mathcal{R}_{Q_{T}}(h_{T,k}^{\star}) + \mathcal{R}_{Q_{T}}(h_{T,k}^{\star}, \hat{h}_{\alpha}).
\end{equation*}
Now, we add and subtract $\mathcal{R}_{P_{k}}(h_{T,k}^{\star}, \hat{h}_{\alpha})$,
\begin{align*}
    \mathcal{R}_{Q_{T}}(\hat{h}_{\alpha}) &\leq (\mathcal{R}_{Q_{T}}(h_{T,k}^{\star},\hat{h}_{\alpha}) - \mathcal{R}_{P_{k}}(h_{T,k}^{\star}, \hat{h}_{\alpha})) + (\mathcal{R}_{Q_{T}}(h_{T,k}^{\star}) + \underbrace{\mathcal{R}_{P_{k}}(h_{T,k}^{\star}, \hat{h}_{\alpha})}_{\leq \mathcal{R}_{P_{k}}(h_{T,k}^{\star}) + \mathcal{R}_{P_{k}}(\hat{h}_{\alpha})}),\\
    &\leq \underbrace{(\mathcal{R}_{Q_{T}}(h_{T,k}^{\star},\hat{h}_{\alpha}) - \mathcal{R}_{P_{k}}(h_{T,k}^{\star}, \hat{h}_{\alpha}))}_{\leq W_{2}(P_{k},Q_{T})\text{ by Lemma 1.}} + \underbrace{(\mathcal{R}_{Q_{T}}(h_{T,k}^{\star}) + \mathcal{R}_{P_{k}}(h_{T,k}^{\star}))}_{=\lambda_{k}\text{ by Def.}} + \mathcal{R}_{P_{k}}(\hat{h}_{\alpha})
\end{align*}
where, in the first line, we used the triangle inequality again. This result leads to,
\begin{equation*}
    \mathcal{R}_{Q_{T}}(\hat{h}_{\alpha})
    \leq \mathcal{R}_{P_{k}}(\hat{h}_{\alpha}) + W_{2}(P_{k}, Q_{T}) + \lambda_{k}.
\end{equation*}
Now, summing over $k$, weighted by $\alpha$, one has $\mathcal{R}_{Q_{T}}(\hat{h}_{\alpha}) = \sum_{k}\alpha_{k}\mathcal{R}_{Q_{T}}(\hat{h}_{\alpha})$. We can bound this latter term as follows,
\begin{align}
    \sum_{k}\alpha_{k}\mathcal{R}_{Q_{T}}(\hat{h}_{\alpha}) &\leq \sum_{k}\alpha_{k}\mathcal{R}_{P_{k}}(\hat{h}_{\alpha}) + \sum_{k}\alpha_{k}(W_{2}(P_{k},Q_{T})+\lambda_{k}),\nonumber\\
    &= \mathcal{R}_{\alpha}(\hat{h}_{\alpha}) + \sum_{k}\alpha_{k}(W_{2}(P_{k},Q_{T})+\lambda_{k}),\nonumber\\
    &\leq \mathcal{R}_{\alpha}(\hat{h}_{\alpha}) + \sum_{k}\alpha_{k}(W_{2}(\hat{P}_{k},\hat{Q}_{T})+\lambda_{k}+\zeta_{k}),\label{eq:intermediate_term_bound_thm2}
\end{align}
From this last inequality, we use the triangle inequality between $\hat{P}_{k}$, $\mathcal{B}(\alpha;\mathcal{P})$ and $\hat{Q}_{T}$,
\begin{align*}
    W_{2}(\hat{P}_{k}, \hat{Q}_{T}) \leq W_{2}(\hat{P}_{k}, \mathcal{B}(\alpha;\mathcal{P}))+ W_{2}(\mathcal{B}(\alpha;\mathcal{P}), \hat{Q}_{T}).
\end{align*}
Summing over $k$, and noting that,
\begin{align*}
    \sum_{k}\alpha_{k}W_{2}(\mathcal{B}(\alpha;\mathcal{P}), \hat{Q}_{T})=W_{2}(\mathcal{B}(\alpha;\mathcal{P}), \hat{Q}_{T})
\end{align*}
one has,
\begin{align*}
    \sum_{k}\alpha_{k}W_{2}(\hat{P}_{k},\hat{Q}_{T}) \leq W_{2}(\mathcal{B}(\alpha;\mathcal{P}), \hat{Q}_{T}) + \gamma.
\end{align*}
Plugging this result back into equation~\ref{eq:intermediate_term_bound_thm2},
\begin{align*}
    \mathcal{R}_{Q_{T}}(\hat{h}_{\alpha}) 
 \leq \mathcal{R}_{\alpha}(\hat{h}_{\alpha}) + W_{2}(\hat{Q}_{T}, \mathcal{B}(\alpha;\mathcal{P})) + \gamma + \lambda + \zeta,
\end{align*}
where,
\begin{align*}
    \gamma = \sum_{k=1}^{K}\alpha_{k}W_{2}(\hat{P}_{k},\mathcal{B}(\alpha;\mathcal{P}))\text{, and, } \lambda = \sum_{k=1}^{K}\alpha_{k}\lambda_{k}\text{, }\xi = \sum_{k=1}^{K}\alpha_{k}\xi_{k}
\end{align*}

which concludes the proof.

This theorem is similar to previous theoretical guarantees for \gls{msda}, such as those of \cite{ben2010theory}, \cite{redko2017theoretical}, and \cite{montesuma2021cvpr}. A few remarks may be made about the terms $\gamma$, $\lambda$ and $\zeta$,
\begin{itemize}
    \item $\zeta$ corresponds to the sample complexity of approximating $W_{2}(P_{k},Q_{T})$ by $W_{2}(\hat{P}_{k},\hat{Q}_{T})$. It decreases with the number of samples in the atoms, and in the target domain.
    \item $\lambda$ corresponds to the domain adaptation complexity between the atoms and the target. As in \gls{dadil}-R, this term cannot be bounded, but it is in general assumed to remain bounded during \gls{da}.
    \item $\gamma$ is a new term, corresponding to $\sum_{k}\alpha_{k}W_{2}(\hat{P_{k}}, \mathcal{B}(\alpha;\mathcal{P}))$. For a given $\alpha$, this term is minimal since $\mathcal{B}(\alpha;\mathcal{P})$ is, by definition, the minimizer of $\hat{B} \mapsto \sum_{k}\alpha_{k}W_{2}(\hat{P}_{k},\hat{B})$.
\end{itemize}

\section{Wasserstein Barycentric Coordinate Regression}\label{appx:wbr}

In this section we describe how to adapt the \gls{wbr} algorithm~\cite{bonneel2016wasserstein} for empirical distributions. Let $\mathcal{Q}_{S} = \{\hat{Q}_{\ell}\}_{\ell=1}^{N_{S}}$ be the set of labeled source distributions, and $\alpha \in \Delta_{N_{S}}$ be a set of \emph{barycentric coordinates}, i.e., they allow to interpolate between distributions in $\mathcal{Q}$ through $\mathcal{B}(\alpha;\mathcal{Q}_{S})$. Let $\hat{Q}_{T}$ be an unlabeled target distirbution. ~\cite{bonneel2016wasserstein} proposed regressing $\alpha$ by the following minimization problem,
\begin{align}
    \alpha^{\star} = \argmin{\alpha \in \Delta_{N_{S}}} W_{2}(\hat{Q}_{T}, \mathcal{B}(\alpha;\mathcal{Q}_{S})).\label{eq:wbr}
\end{align}
In~\cite{bonneel2016wasserstein}, the minimization in equation~\ref{eq:wbr} was done when $\hat{Q}_{S_{\ell}}$ and $\hat{Q}_{T}$ are histograms, i.e., vectors in $\Delta_{d}$. In our case, these distributions are empirical. As a consequence, we can employ a similar method to \gls{dadil} for regressing $\alpha \in \Delta_{N_{S}}$, as shown in algorithm~\ref{alg:wbr}.

\begin{algorithm}[H]
    \caption{Wasserstein Barycentric Regression of Point Clouds.}
    \begin{algorithmic}[1]
        \small
        \REQUIRE $\mathcal{Q}_{S} = \{\hat{Q}_{S_{\ell}}\}_{\ell=1}^{N_{S}}$, $\hat{Q}_{T}$, batch size $n_{b}$, learning rate $\eta$, number of iterations $N_{iter}$, number of batches $M$.
        \STATE Initialize $\alpha_{\ell} = N_{S}^{-1}$, $\ell=1,\cdots,N_{S}$.
        \FOR{$it=1\cdots,N_{iter}$}
            \FOR{$batch=1,\cdots,M$}
                \FOR{$\ell=1,\cdots,N_{S}$}
                    \STATE Sample $\mathbf{X}^{(Q_{S_{\ell}})} = \{\mathbf{x}_{i}^{(Q_{S_{\ell}})}\}_{i=1}^{n_{b}}$
                \ENDFOR
                \STATE sample $\mathbf{X}^{(Q_{T})} = \{\mathbf{x}_{j}^{(Q_{T})}\}_{j=1}^{n_{b}}$
                \STATE calculate $\mathbf{X}^{(B)} = \mathcal{B}(\alpha_{\ell};\mathcal{Q}_{S})$
                \STATE $L \leftarrow W_{2}(\hat{Q}_{T}, \hat{B})$
                \STATE $\alpha \leftarrow \text{proj}_{\Delta_{N_{S}}}(\alpha - \eta \nicefrac{\partial L}{\partial \alpha_{\ell}})$.
            \ENDFOR
        \ENDFOR
        \ENSURE Barycentric coordinates $\alpha^{\star}$.
    \end{algorithmic}
    \label{alg:wbr}
\end{algorithm}

As noted by~\cite{bonneel2016wasserstein}, the barycentric regression algorithm leads to poor reconstructions, when $\hat{Q}_{T}$ is far from the distributions in $\mathcal{Q}_{S}$. This remark explains its sub-optimal performance in our ablation study (i.e., table 3 of our main paper). These results also support why \gls{dil} is important in \gls{msda}, as learning atom distributions leads to a Wasserstein hull $\mathcal{P}$ for which $\hat{Q}_{T}$ is close.

\section{Experiments Details}\label{appx:experiments}

\subsection{Dataset Description}

\noindent\textbf{Caltech-Office 10:} We use the experimental setting of~\cite{montesuma2021cvpr}, namely, the 5 fold cross-validation partitions and the features (DeCAF 7th layer activations).

\noindent\textbf{Refurbished-Office 31:} This dataset consists on the Office 31 dataset of~\cite{saenko2010adapting}. We follow the discussion of~\cite{ringwald2021adaptiope} and adopt their modifications, namely, the replacements of images in the Amazon domain that contained label noise. The authors modifications are publicly available in a Gitlab respository\footnote{\url{https://gitlab.com/tringwald/adaptiope}}. Since our method relies on features, we adopt the following methodology,
\begin{enumerate}
    \item A backbone deep \gls{nn} is fitted using all source data. For instance if adaptation is done between $A,D\rightarrow W$, then all data from $A$ and $D$ is used for training the \gls{nn}.
    \item Features are extracted using the convolutional layers. This generates a new dataset of shape $(n_{samples}, d)$, where $d$ is the dimensionality of features.
\end{enumerate}
We split each domain in a train and test set, corresponding to 80\% and 20\% of samples, respectively. This is shown in Table~\ref{tab:da_info}. Here, we highlight that during fine-tuning in step $1$, all data from sources is used (i.e., training and test partitions), as evaluation is performed solely on target data. \gls{dadil} is then run using the features extracted at step $2$. For the Refurbished-Office 31, we chose to use the ResNet-50 of~\cite{he2016deep} as backbone.

Since this benchmark has a significant performance saturation (i.e., fine-tuning a complete ResNet yields 100\% classification accuracy when adapting towards \emph{dSLR} and \emph{Webcam}), we use a slightly more challenging setting, where we keep all but the last convolutional block frozen during fine-tuning.

\noindent\textbf{CWRU:} The third benchmark we experiment on is the Case Western Reserve University bearing fault diagnosis benchmark, which is publicly available\footnote{\url{https://engineering.case.edu/bearingdatacenter/download-data-file}}. We use the 12k drive end bearing fault data, where we consider the file ids reported in table~\ref{tab:fileids_crwu}. This is similar to other studies in bearing fault diagnosis, as reported in~\cite{zhang2018intelligent}.

\begin{table}[ht]
    \centering
    \caption{File ids of data used in the CWRU experiments.}
    \begin{tabular}{ccccccccccc}
        \toprule
        Label & 0 & 1 & 2 & 3 & 4 & 5 & 6 & 7 & 8 & 9 \\
        Fault Location & None & \multicolumn{3}{c}{Inner Race Fault} & \multicolumn{3}{c}{Ball Fault} & \multicolumn{3}{c}{Outer Race Fault}\\
        Fault Diameter & & 0.007 & 0.014 & 0.021 & 0.007 & 0.014 & 0.021 & 0.007 & 0.014 & 0.021\\
        \midrule
        1772 rpm & 98  & 106 & 170 & 210 & 119 & 186 & 223 & 131 & 198 & 235\\
        1750 rpm & 99  & 107 & 171 & 211 & 120 & 187 & 224 & 132 & 199 & 236\\
        1730 rpm & 100 & 108 & 172 & 212 & 121 & 188 & 225 & 133 & 200 & 237 \\
        \bottomrule
    \end{tabular}
    \label{tab:fileids_crwu}
\end{table}

In the CWRU benchmark, for each file id we randomly sample a window of length 4096 from the complete signal. From this sub-sample, we extract the first 2048 Fourier coefficients, from the 4096 acquires through a Fast Fourier Transform (FFT). For each target domain, we pre-train a multi-layer Perceptron\footnote{$2048 \rightarrow 1024 \rightarrow 512 \rightarrow 256$ with ReLU activations.} from scratch, and use the activation of its penultimate layer as features for shallow methods, including \gls{dadil}.

Overall, when training \gls{dadil}, we use all source data (i.e. train and test), and the train partition of the target domain, without its labels. The performance evaluation is done in the independent test set, which is never seen during training.

\begin{table}[ht]
    \centering
    \caption{Details about the datasets considered for domain adaptation.}
    \begin{tabular}{cccccc}
        \toprule
        Dataset & \# Classes & Domain & \# Training Samples & \# Test Samples & \# Samples\\
        \midrule
        \multirow{4}{*}{Caltech-Office 10} & \multirow{4}{*}{10} & Amazon & 748 & 210 & 958\\
        & & dSLR & 108 & 49 & 157\\
        & & Webcam & 224 & 71 & 295\\
        & & Caltech & 956 & 224 & 1180\\
        & & Total & 2036 & 554 & 2590\\
        \midrule
        \multirow{4}{*}{Office 31} & \multirow{4}{*}{31} & Amazon &  2253 & 564 & 2817\\
                & & dSLR & 398 & 100 & 498\\
                & & Webcam & 636 & 159 & 795\\
        &       & Total & 3287 & 823 & 4110\\
        \midrule
        \multirow{4}{*}{CWRU} & \multirow{4}{*}{10} & 1772rpm & 6000 & 2000 & 8000\\
                              &  & 1750rpm & 6000 & 2000 & 8000\\
                              &  & 1720rpm & 6000 & 2000 & 8000\\
                              & & Total & 18000 & 6000 & 24000\\
        \bottomrule
    \end{tabular}
    \label{tab:da_info}
\end{table}

\subsection{DaDiL Implementation Details}

In this section we highlight a few details in our implementation. Concerning algorithm 1, we remark that we differentiate $\mathbf{X}^{(B)}$ and $\mathbf{Y}^{(B)}$ \emph{at termination} of algorithm 1 (main paper). This means that derivatives are not back-propagated through the while loop in algorithm 1 (main paper). In practical terms, this corresponds to stating \emph{with torch.no\_grad()} before the loop, then using the transport plans $\pi^{(k)}$ at termination for calculating the support $\mathbf{X}^{(B)}$, $\mathbf{Y}^{(B)}$. This is similar to what was proposed by~\cite{feydy2019interpolating}. In practice, $\pi^{(k)}_{i,l}$ is not differentiated w.r.t. $\mathbf{x}_{l}^{(P_{k})}, \mathbf{y}_{l}^{(P_{k})}$ or $\alpha$.

We also implement algorithm 2 using Pytorch, thus calculating the derivatives of $L(\mathcal{P},\mathcal{A})$ automatically. The projection into the simplex is done using \gls{pot}'s function \emph{ot.utils.proj\_simplex}, which does so by sorting (see e.g.,~\cite{held1974validation} and~\cite{condat2016fast} for more details).

\section{Additional Experiments}\label{appx:additional_exps}

In this section, we provide additional experiments. In section~\ref{sec:hyperparam_analysis}, we explore the reasoning for tuning the hyper-parameters of \gls{dadil}. In addition, section~\ref{sec:learning-loop-stability} we explore the stability of algorithm 2 w.r.t. different random initializations of atoms, as well as the evolution of weights $\alpha$ w.r.t. iterations.

\subsection{Hyper-parameter Analysis}\label{sec:hyperparam_analysis}

In our experiments, we tune hyper-parameters so as to maximize performance on target domain training data. Our evaluation is done in an independent partition of the test set, not seen during training. In what follows, we highlight that previous best-performing method in the \gls{sota}, i.e. WBT$_{reg}$, has $92.55\%$ of average domain performance.

\noindent\textbf{Number of Atoms.} We test $K \in \{3, \cdots, 8\}$. We divide our discussion into two experiments: (i) \gls{da} performance, and (ii) weight sparsity as a function of the number of atoms $K$. We fix $n = 1000$ and $n_{b} = 200$. First, we measure classification performance on the target domain, which is shown in Table~\ref{tab:variable_K}. Performance fluctuates over choices of $K$, being higher with increasing number of components. For all choices, we improve over previous \gls{sota}.

\begin{table}[ht]
    \centering
    \caption{Classification accuracy in \%, as a function of number of components $K$.}
    \begin{tabular}{ccccccc}
        \toprule
        $K$ & 3 & 4 & 5 & 6 & 7 & 8\\ 
        \midrule
        DaDiL-R & 93.53 & 94.16 & 93.33 & 94.30 & 94.23 & 95.30\\ 
        DaDiL-E & 93.89 & 94.33 & 93.86 & 94.68 & 94.66 & 95.70\\
        \bottomrule
    \end{tabular}
    \label{tab:variable_K}
\end{table}

Second, for a fixed $K$ we measure the sparsity of $\alpha_{T}$ through the following score,
\begin{equation*}
    \text{sparsity score} = 100\% \times \biggr(1-\dfrac{\lVert \alpha_{T} \rVert_{0}}{K}\biggr),
\end{equation*}
that is, the percentage of zero entries in the target domain weights.

\begin{table}[ht]
    \centering
    \caption{Sparsity Score (in \%) as a function of number of atoms $K$.}
    \begin{tabular}{ccccccccc}
        \toprule
        $K$ & 3 & 4 & 5 & 6 & 7 & 8 \\ 
        \midrule
        Amazon & 6.66 & 0.00 & 8.00 & 20.00 & 8.58 & 15.00\\ 
        Caltech & 6.66 & 20.00 & 16.00 & 30.00 & 31.42 & 27.50 \\ 
        dSLR & 0.00 & 5.00 & 8.00 & 23.33 & 25.71 & 27.50 \\ 
        Webcam & 6.66 & 5.00 & 12.00 & 10.00 & 20.00 & 27.50 \\ 
        Avg & 5.00 & 7.50 & 11.00 & 20.83 & 21.43 & 24.37 \\ 
        \bottomrule
    \end{tabular}
    \label{tab:atom_sparsity}
\end{table}

As the number of atoms grows, $\alpha_{T}$ becomes more sparse. This implies that \gls{dadil} selects a sub-set of atoms to reconstruct the target domain whenever $K$ is overestimated.

\noindent\textbf{Number of Samples.} We divide our discussion into two experiments: (i) \gls{da} performance and (ii) support density, as a function of the number of samples $n$ in the support of $\hat{P}_{k}$. We fix $K = 8$ and $n_{b} = 200$. First, as shown in table~\ref{tab:perf_n_samples}, \gls{dadil} is more sensible to the number of samples in the support of atoms. While the best choice is $n = 1000$, for other values of $n$ we remain \gls{sota}.

\begin{table}[ht]
    \centering
    \caption{Classification accuracy in \%, as a function of the number of samples $n$.}
    \begin{tabular}{ccccc}
        \toprule
        $n$ & 500 & 1000 & 2000 \\ 
        \midrule
        DaDiL-R & 92.48 & 95.30 & 93.57 \\ 
        DaDiL-E & 93.12 & 95.70 & 93.58 \\ 
        \bottomrule
    \end{tabular}
    \label{tab:perf_n_samples}
\end{table}

Second, let us explore the density of samples in the atoms support. Let $N_{5}(\mathbf{x}_{i}^{(P_{k})})$ denote the set of 5 nearest neighbors in $\hat{P}_{k}$, w.r.t. $\mathbf{x}_{i}^{(P_{K})}$. We evaluate the density of atoms support as a function of $n$ by measuring the average distance of each point in $\mathbf{X}^{(P_{k})}$ to its $5$ nearest neighbors, i.e.,
\begin{equation*}
    \text{density score} = \dfrac{1}{5nK}\sum_{k=1}^{K}\sum_{i=1}^{n}\sum_{\mathbf{x}_{j}^{(P_{k})}\in N_{5}(\mathbf{x}_{i}^{(P_{k})})}\lVert \mathbf{x}_{i}^{(P_{k})} - \mathbf{x}_{j}^{(P_{k})}\rVert_{2}^{2}.
\end{equation*}

As table~\ref{tab:atom_density} shows, the samples concentrate over a region of $\mathbb{R}^{d}$, meaning that the atoms' support are stable w.r.t $n$.

\begin{table}[ht]
    \centering
    \caption{Average distance to 5 nearest neighbors of points in the learned atoms support.}
    \begin{tabular}{ccccc}
        \toprule
        $n$ & 1000 & 2000 & 3000 & 4000 \\ 
        \midrule
        Amazon & 43.28 $\pm$ 0.42 & 34.92 $\pm$ 1.14 & 33.04 $\pm$ 1.80 & 32.18 $\pm$ 1.45 \\ 
        Caltech & 43.12 $\pm$ 0.60 & 34.49 $\pm$ 1.57 & 34.33 $\pm$ 2.04 & 32.97 $\pm$ 2.28 \\ 
        dSLR & 43.60 $\pm$ 0.96 & 35.82 $\pm$ 0.39 & 33.03 $\pm$ 0.47 & 32.72 $\pm$ 0.70 \\ 
        Webcam & 42.71 $\pm$ 1.01 & 36.07 $\pm$ 0.93 & 35.14 $\pm$ 2.69 & 33.45 $\pm$ 1.00 \\ 
        Avg & 43.18 & 35.31 & 33.88 & 32.83 \\ 
        \bottomrule
    \end{tabular}
    \label{tab:atom_density}
\end{table}

\noindent\textbf{Batch size $n_{b}$.} For fixed $K = 8$ and $n = 1000$, we evaluate $n_{b} \in \{50, 100, 200\}$ in terms of average domain performance. Our results are shown in Table~\ref{tab:perf_batch_size}. Note that performance increases with an increasing batch size, which agrees with previous research in mini-batch \gls{ot}~\cite{fatras2020learning,fatras2021minibatch}. In addition, performance is more stable w.r.t. batch size choice (e.g., compare tables~\ref{tab:perf_batch_size} and~\ref{tab:perf_n_samples}), and \gls{dadil} remains largely above previous \gls{sota} for all choices of batch size.

\begin{table}[ht]
    \centering
    \caption{Classification accuracy in \%, as a function of batch size $n_{b}$.}
    \begin{tabular}{cccc}
        \toprule
        $n_{b}$ & 50 & 100 & 200 \\ 
        \midrule
        DaDiL-R & 94.78 & 94.94 & 95.70 \\ 
        DaDiL-E & 94.64 & 94.72 & 95.30  \\ 
        \bottomrule
    \end{tabular}
    \label{tab:perf_batch_size}
\end{table}

\noindent\textbf{Learning rate and Number of iterations.} Since we use large, balanced batches, our optimization problem is stable w.r.t. learning rate and number of iterations. Overall, we set $\eta$ as the maximum value for which our algorithm does not diverge. \gls{dadil} is run until reconstruction loss hits a plateau. Note that this criteria does not assumes labeled samples from the target domain.

\noindent\textbf{Final Hyper-parameter Settings.} Overall, we select hyper-parameters that maximize classification performance on target domain samples seen during training. For evaluation, we use an independent partition of the target domain data not seen during training. We report the used hyper-parameters in table~\ref{tab:hyperparam_info}.

\begin{table}[ht]
    \centering
    \caption{Final hyper-parameter settings for each benchmark.}
    \begin{tabular}{cccccc}
        \toprule
        Dataset & & DaDiL-R & DaDiL-E\\
        \midrule
        \multirow{5}{*}{Caltech-Office 10} & $K$ & 8 & 8 \\
        & $n$ & 1000 & 1000\\
        & $n_{b}$ & 200 & 200\\
        & $\eta$ & $10^{-1}$ & $10^{-1}$ \\
        & $N_{iter}$ & 10 & 10\\
        \midrule
        \multirow{5}{*}{Refurbished Office 31} & $K$ & 6 & 6\\
        & $n$ & 3100 & 1550\\
        & $n_{b}$ & 155 & 310\\
        & $\eta$ & $5 \times 10^{-2}$ & $5 \times 10^{-2}$ \\
        & $N_{iter}$ & 60 & 60\\
        \midrule
        \multirow{5}{*}{CWRU} & $K$ & 8 & 3 \\
        & $n$ & 1000 & 1000 \\
        & $n_{b}$ & 200 & 200 \\
        & $\eta$ & $5 \times 10^{-2}$ & $5 \times 10^{-2}$ \\
        & $N_{iter}$ & 60 & 60\\
        \bottomrule
    \end{tabular}
    \label{tab:hyperparam_info}
\end{table}

\subsection{Learning Loop Stability}\label{sec:learning-loop-stability}

In this section, we analyze the stability of \gls{dadil}'s learning problem (i.e., the minimization of eqn. 11) under different initializations of atom's support and weights. This experiment is conducted on the CWRU benchmark, where we evaluate stability w.r.t. 2 criteria,
\begin{enumerate}
    \item The stability of the loss curve w.r.t. different and independent initializations,
    \item The stability of parameter updates w.r.t. iterations,
\end{enumerate}

For the first criterion, we calculate the mean and standard deviation of $L_{it}$ over 5 independent runs. Overall, the loss is stable w.r.t. random initializations, and it converges to different local minima, as shown by the small variation in the loss at the beginning and end of training.

\begin{figure}[ht]
    \centering
    \includegraphics[width=0.8\linewidth]{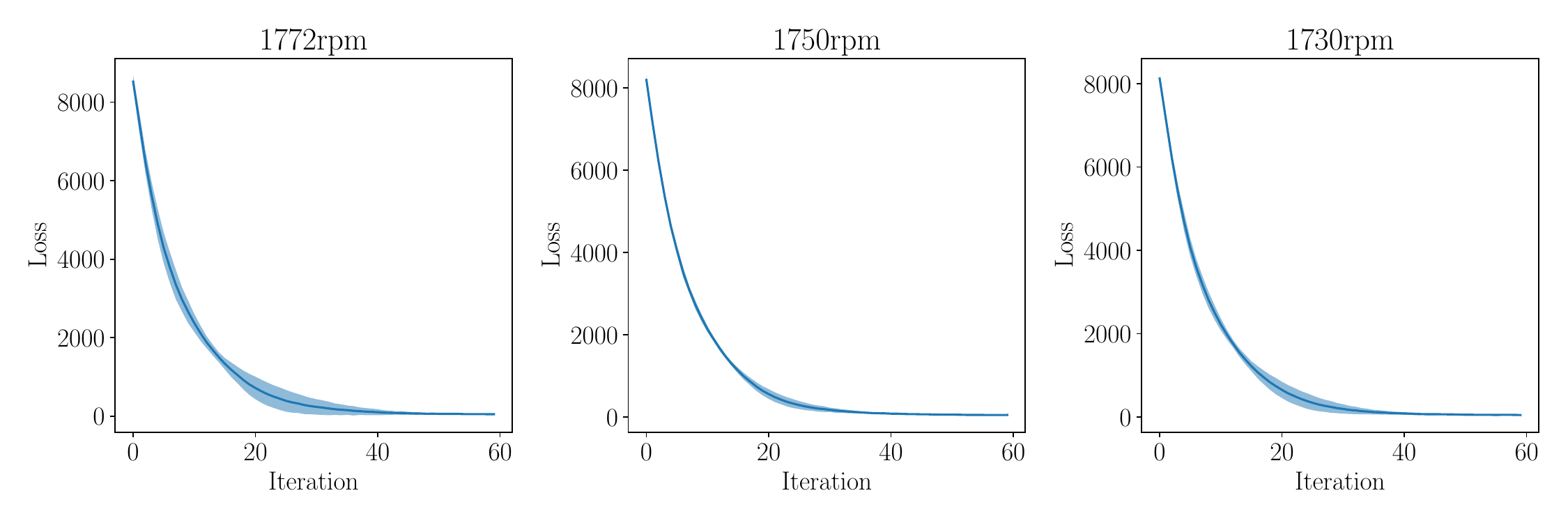}
    \caption{\gls{dadil}'s loss function (eqn 11) as a function of iterations. The solid line represents an average over 5 independents runs, whereas the shaded region represents a confidence interval of $\pm 2\sigma$.}
    \label{fig:loss_stability}
\end{figure}

These experiments raise the question on whether minimizers to eqn. 11 are unique. It is generally not the case. For instance, if one wants to express $\mathcal{Q} = \{Q_{0}\}$, $Q_{0} = \mathcal{N}(\mu, I)$ as the barycenter of $\mathcal{P} = \{P_{1},P_{2}, P_{3}\}$, where $P_{i} = \mathcal{N}(\mu_{i},I)$, any set of means with $(\nicefrac{1}{3})\sum_{i=1}^{3}\mu_{i} = \mu$ is a minimizer of eqn. 11. Thus theoretical analysis of minimizers of eqn. 11 is challenging, and we will consider it in future works.

Concerning the second criterion, we evaluate the magnitude of updates in $\mathbf{X}^{(P_{k})}_{it}$, $\mathbf{Y}^{(P_{k})}_{it}$ and $\mathcal{A}_{it}$. We use the Frobenius norm for comparing updates, i.e.,
\begin{equation}
    \begin{aligned}
        \Delta_{X}(it) &= \dfrac{1}{K}\sum_{k=1}^{K}\lVert \mathbf{X}_{it}^{(P_{k})} - \mathbf{X}_{it-1}^{(P_{k})}\rVert_{F}^{2},\\
        \Delta_{Y}(it) &= \dfrac{1}{K}\sum_{k=1}^{K}\lVert \mathbf{Y}_{it}^{(P_{k})} - \mathbf{Y}_{it-1}^{(P_{k})}\rVert_{F}^{2},\\
        \Delta_{A}(it) &= \lVert \mathcal{A}_{it} - \mathcal{A}_{it-1}\rVert_{F}^{2}.
    \end{aligned}\label{eq:magnitudes}
\end{equation}

\begin{figure}[ht]
    \centering
    \includegraphics[width=0.8\linewidth]{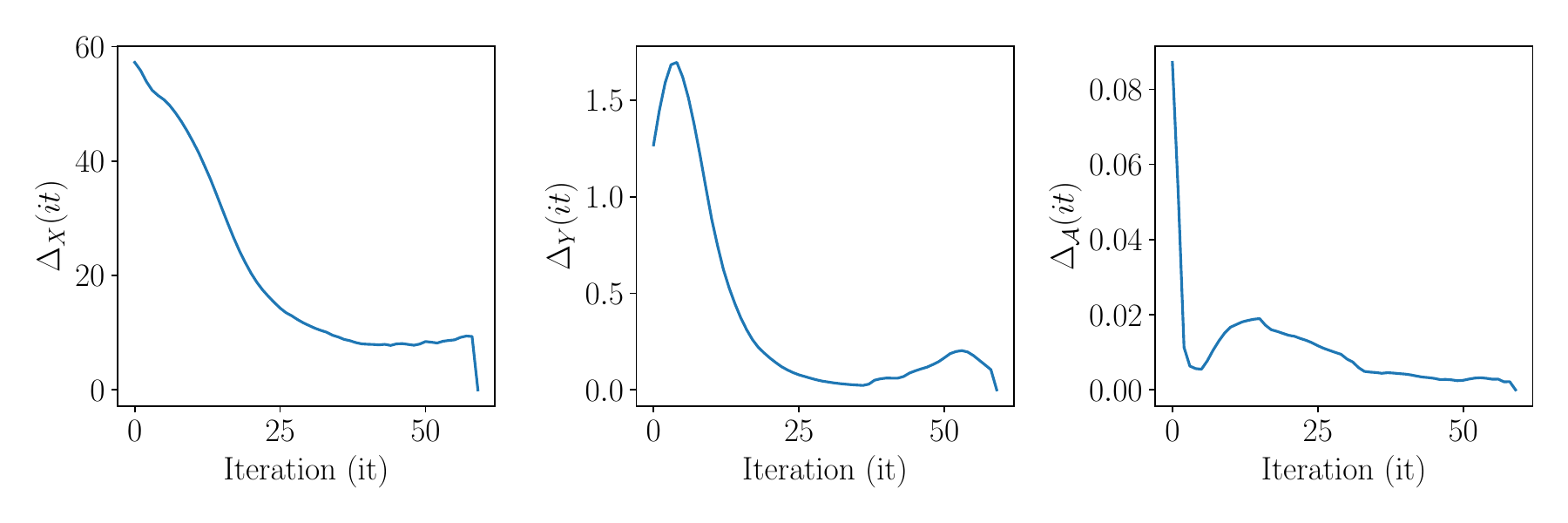}
    \caption{Magnitude of updates of \gls{dadil}'s algorithm as a function of iterations.}
    \label{fig:stability_variables}
\end{figure}

We show our results in figure~\ref{fig:stability_variables}. Note that, towards the end of learning, the updates on \gls{dadil}'s variables fall close to 0, indicating that the algorithm converged to a local minimum.

\end{document}